\documentclass{article}

\usepackage{microtype}
\usepackage{graphicx}
\usepackage{booktabs} 

\usepackage{amsmath,amsfonts,bm}

\def\eqref#1{equation~\ref{#1}}

\def\1{\bm{1}}

\DeclareMathAlphabet{\mathsfit}{\encodingdefault}{\sfdefault}{m}{sl}
\SetMathAlphabet{\mathsfit}{bold}{\encodingdefault}{\sfdefault}{bx}{n}

\usepackage{hyperref}
\usepackage{xurl}  
\usepackage{multirow}
\usepackage{xspace}
\usepackage[disable]{todonotes}
\usepackage{enumitem} 
\usepackage{subcaption}
\usepackage{xcolor}
\usepackage{listings}
\usepackage[most]{tcolorbox}
\usepackage{tikz}
\usetikzlibrary{shadows}
\tcbuselibrary{listings,skins,breakable}
\usepackage{threeparttable}  
\usepackage{cleveref}

\usepackage[normalem]{ulem} 

\usepackage[accepted]{mlsys2025}

\newcommand{\ours}{\textsc{ADR}\xspace}
\newcommand{\company}{\textsc{Uber}\xspace}

\newcommand{\cameraready}[1]{#1}

\mlsystitlerunning{\ours: An Agentic Detection System for Enterprise Agentic AI Security}

\definecolor{adsBG}{HTML}{F8F9FB}
\definecolor{adsBorder}{HTML}{DDE1E6}
\definecolor{adsKeyword}{HTML}{005CC5}
\definecolor{adsComment}{HTML}{6A737D}
\definecolor{adsString}{HTML}{22863A}
\definecolor{adsEASBG}{HTML}{E8F5E9} 
\definecolor{adsEASBorder}{HTML}{A5D6A7}
\definecolor{adsEASShadow}{HTML}{C8E6C9}
\definecolor{adsCurBG}{HTML}{E3F2FD} 
\definecolor{adsCurBorder}{HTML}{90CAF9}

\lstdefinestyle{adsPython}{
  language=Python,
  basicstyle=\ttfamily\footnotesize,
  keywordstyle=\color{adsKeyword}\bfseries,
  commentstyle=\color{adsComment}\itshape,
  stringstyle=\color{adsString},
  upquote=true,
  showstringspaces=false,
  breaklines=true,
  breakatwhitespace=true,
  columns=fullflexible,
  keepspaces=true
}

\lstdefinestyle{adsYAML}{
  basicstyle=\ttfamily\footnotesize,
  commentstyle=\color{adsComment}\itshape,
  showstringspaces=false,
  breaklines=true,
  breakatwhitespace=true,
  columns=fullflexible,
  keepspaces=true,
  literate={[EAS]}{{\adsEASTag{EAS}}}{5}
           {[CURATED]}{{\adsCurTag{CURATED}}}{9}
}

\newtcblisting{AdsCodePy}[1][]{listing only,listing options={style=adsPython,aboveskip=0mm,belowskip=0mm,breaklines=true,breakatwhitespace=true},colback=adsBG,colframe=adsBorder,boxrule=0.4pt,arc=1mm,left=2mm,right=2mm,top=1mm,bottom=1mm,#1}
\newtcblisting{AdsCodeYAML}[1][]{listing only,listing options={style=adsYAML,aboveskip=0mm,belowskip=0mm,breaklines=true,breakatwhitespace=true},colback=adsBG,colframe=adsBorder,boxrule=0.4pt,arc=1mm,left=2mm,right=2mm,top=1mm,bottom=1mm,#1}

\newcommand{\adsEASTag}[1]{\tcbox[on line,colback=adsEASBG,colframe=adsEASBorder,boxrule=0pt,arc=0.9mm,enhanced,drop shadow={shadow xshift=0.6mm,shadow yshift=-0.6mm,opacity=0.25},left=1.5pt,right=1.5pt,top=0.2pt,bottom=0.2pt]{#1}}
\newcommand{\adsCurTag}[1]{\tcbox[on line,colback=adsCurBG,colframe=adsCurBorder,boxrule=0pt,arc=0.9mm,left=1.5pt,right=1.5pt,top=0.2pt,bottom=0.2pt]{#1}}

\begin{document}

\twocolumn[
\mlsystitle{\ours: AN AGENTIC DETECTION SYSTEM FOR ENTERPRISE AGENTIC AI SECURITY}



\mlsyssetsymbol{equal}{*}

\begin{mlsysauthorlist}
\mlsysauthor{Chenning Li}{uber,mit}
\mlsysauthor{Pan Hu}{uber}
\mlsysauthor{Justin Xu}{uber,oxford}
\mlsysauthor{Baris Ozbas}{uber}
\mlsysauthor{Olivia Liu}{uber}
\mlsysauthor{Caroline Van}{uber}
\mlsysauthor{Manxue Li}{uber}
\mlsysauthor{Wei Zhou}{uber}
\mlsysauthor{Mohammad Alizadeh}{mit}
\mlsysauthor{Pengyu Zhang}{uber}
\mlsysauthor{KK Sriramadhesikan}{uber}
\mlsysauthor{Ming Zhang}{uber}
\end{mlsysauthorlist}

\mlsysaffiliation{uber}{Uber Technologies, Inc., USA}
\mlsysaffiliation{mit}{Massachusetts Institute of Technology, USA}
\mlsysaffiliation{oxford}{University of Oxford, United Kingdom}

\mlsyscorrespondingauthor{Pan Hu}{lghupan@gmail.com}

\mlsyskeywords{AI Agent Security, Model Context Protocol, Agentic AI, Enterprise Security, Red Teaming, Benchmark}

\vskip 0.3in

\begin{abstract}

We present the \textbf{Agentic AI Detection and Response (ADR)} system, the first \textit{large-scale, production-proven} enterprise framework for securing AI agents operating through the Model Context Protocol (MCP). We identify three persistent challenges in this domain: 
(1) \textbf{limited observability} -- existing \cameraready{Endpoint Detection and Response (EDR)} tools see file writes but not the agent reasoning, prompts, or causal chains linking intent to execution;
(2) \textbf{insufficient robustness} -- static defenses constrained by pre-defined rules fail to generalize across diverse attack techniques and enterprise contexts; and 
(3) \textbf{high detection costs} -- LLM-based inference is prohibitively expensive at scale.
ADR addresses these challenges via three components: the \textbf{\ours Sensor} for high-fidelity agentic telemetry, the \textbf{\ours Explorer} for systematic pre-deployment red teaming and hard-example generation, and the \textbf{\ours Detector} for scalable, two-tier online detection combining fast triage with context-aware reasoning. 
Deployed at \company for over ten months, ADR has sustained reliable detection in production with growing adoption reaching over 7,200 unique hosts and processing over 10,000 agent sessions daily, uncovering hundreds of credential exposures across 26 categories and enabling a shift-left prevention layer (97.2\% precision, 206 detected credentials).
To validate the approach and enable community adoption, we introduce \textbf{\ours-Bench} (302 tasks, 17 techniques, 133 MCP servers), where ADR achieves \textbf{zero false positives} while detecting \textbf{67\% of attacks} -- outperforming three state-of-the-art baselines (ALRPHFS, GuardAgent, LlamaFirewall) by \textbf{2--4$\times$} in F1-score. 
On \textbf{AgentDojo} (public prompt injection benchmark), ADR detects \textbf{all attacks} with only \textbf{three false alarms} out of 93 tasks. 
\end{abstract}

]



\printAffiliationsAndNotice{Work done during internships at Uber Technologies, Inc.}

\section{Introduction}
\label{sec:intro}

The rapid adoption of AI agents capable of autonomous decision-making and tool use is reshaping enterprise workflows.
By using the Model Context Protocol~\cite{modelcontextprotocol-repositories} (MCP) -- a standardized interface for connecting large language models (LLMs) to external tools and data sources -- enterprises now deploy agents that can analyze documents, modify infrastructure, generate code, and interact with internal systems at scale.
While this paradigm unlocks unprecedented efficiency and automation, it also introduces a fundamentally new attack surface: agents can be manipulated through natural language, exploited via compromised MCP servers, or coerced into executing unsafe commands and exfiltrating sensitive data.

Traditional enterprise defenses, such as endpoint detection and response (EDR) tools, static guardrails, and rule-based policy checkers, are ill-suited for this new threat model.
EDR systems can observe file writes and network calls, but cannot see \emph{why} an agent performed those actions, which is captured in the user prompts, agent reasoning steps, or the causal chain linking intent to tool execution.
Static defenses constrained by pre-defined policies struggle to generalize across the diverse landscape of agentic attacks, from prompt injection to tool manipulation to credential exfiltration.
Limited observability, insufficient generalization across attack types, and severe class imbalance make existing mechanisms insufficient for securing AI agents at enterprise scale.

\begin{figure}[t]
\small
\centering
\includegraphics[width=0.45\textwidth]{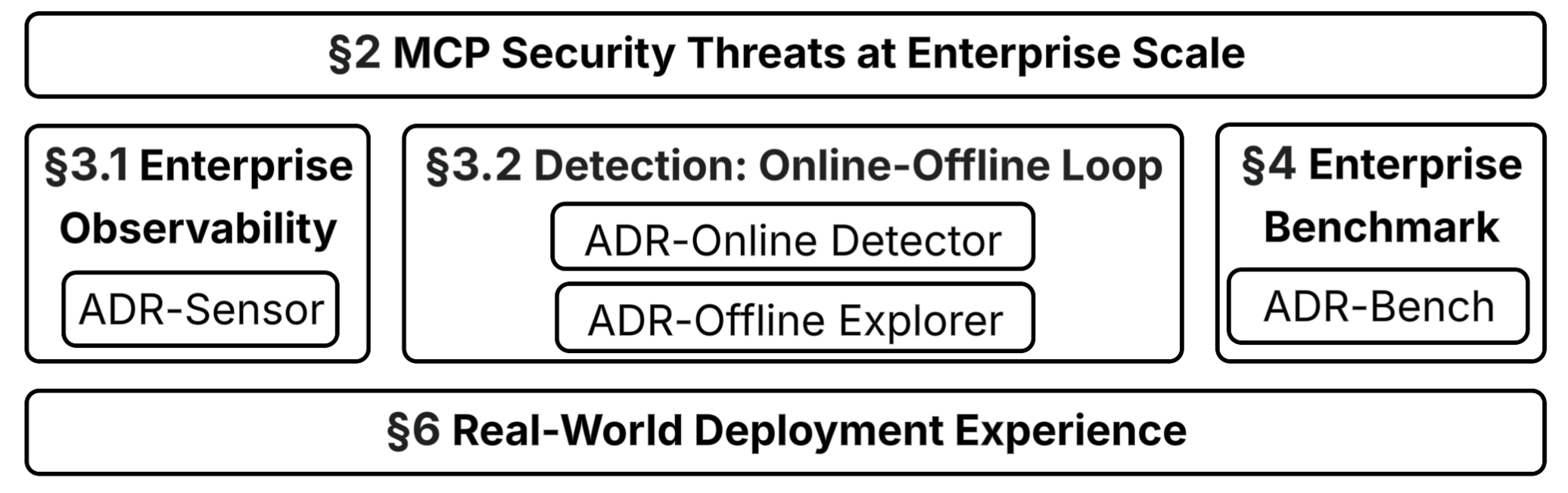}
\caption{\textbf{Enterprise Demands and Our Contributions} for securing agentic AI. 
(1) \textbf{Enterprise Observability} via the \ours Sensor, 
(2) \textbf{Reliable Detection} via the online \ours Detector and offline \ours Explorer, and
(3) \textbf{Enterprise Benchmarking} via \ours-Bench.}
\label{fig:intro_overview}
\end{figure}

This paper presents the \textbf{Agentic AI Detection and Response (\ours)} system, the first end-to-end, enterprise-grade framework for securing MCP-driven AI systems.
As illustrated in \Cref{fig:intro_overview}, \ours is designed to meet three core requirements necessary for safe and scalable agentic operations:
\begin{itemize}[topsep=0pt,itemsep=-1ex,partopsep=1ex,parsep=1ex,leftmargin=*]
    \item \textbf{Enterprise Observability (\S\ref{subsec-observability}).} At the foundation lies the \textbf{\ours Sensor}, a lightweight endpoint component that reconstructs high-fidelity telemetry of agentic workflows.  
    Unlike conventional telemetry systems that capture only outcomes (e.g., file writes or API calls), the \ours Sensor records the full causal chain of prompts, reasoning steps, tool invocations, and environmental context, thus closing the observability gap for AI-driven activity.

    \item \textbf{Reliable, Cost-efficient Detection (\S\ref{subsec-detection-loop}).} At production scale (processing over 10,000 agent sessions daily), LLM-based detection for every event becomes prohibitively expensive. The \textbf{\ours Detector} employs a two-tier architecture that combines fast triage for high recall with deep, context-aware reasoning for precision.  
    To sustain detection robustness across diverse attack types, the offline \textbf{\ours Explorer} engine systematically red-teams the system during pre-deployment testing, discovering hard attack variants and generating threat intelligence that strengthens the detector before production deployment.
    
    \item \textbf{Enterprise Benchmarking (\S\ref{sec-benchmark}).} We introduce \textbf{\ours-Bench} (302 tasks, 42 malicious, 260 benign), a benchmark derived from real enterprise telemetry across 133 MCP servers and 17 attack techniques.  
    It captures the complexity, imbalance, and contextual diversity of production environments, enabling rigorous and reproducible evaluation of agentic security systems.
\end{itemize}

Deployed at \company
for over ten months (\S\ref{subsec-eval-deployment}), \ours has demonstrated sustained reliability with growing adoption reaching over 7,200 unique hosts.
The system detected hundreds of credential exposures across 26 categories that had been inadvertently shared outside the enterprise network.
These findings informed a shift-left prevention layer that achieved 97.2\% precision in blocking credential leaks (206 detected across 212 unique credentials from hundreds of thousands of sessions).
Additionally, controlled testing through internal capture-the-flag exercises and emulation of real-world attacks (Agent Flayer) validated \ours's ability to trace multi-stage prompt injection and exfiltration chains.
To rigorously validate the approach and enable community adoption (\S\ref{sec-evaluation}), we introduce \textbf{\ours-Bench} (302 tasks, 42 malicious, 260 benign) derived from enterprise telemetry across 133 MCP servers and 17 attack techniques.
On \ours-Bench, \ours achieves \textbf{zero false positives} while detecting 67\% of attacks, outperforming baselines by 2--4$\times$ in F1-score while maintaining low latency and cost. We deliberately prioritize precision for production viability, as the high false positive rates of baseline methods (up to 40 FPs out of 260 benign tasks) make them unsuitable for deployment where false alarms trigger expensive incident response.
On AgentDojo~\cite{debenedetti2024agentdojo} (a public prompt injection benchmark), \ours detects all attacks with only three false alarms out of 93 tasks.
These results establish \ours as \textbf{the first \textit{enterprise-proven} framework for securing AI agents and MCP-based systems at scale}.

\cameraready{\ours-Bench and the source code for the \ours Sensor and detection framework are publicly available on \href{https://github.com/uber/ADR}{GitHub}.}
\section{Background \& Motivations}\label{sec-motivation}

\subsection{Background: Model Context Protocol (MCP) and Ecosystem}

The \textit{Model Context Protocol (MCP)}~\cite{modelcontextprotocol-repositories}, introduced by Anthropic in late 2024, defines a standardized interface for AI agents to interact with external tools, systems, and contextual data. 
MCP has rapidly become the backbone of modern agentic workflows across major ecosystems, including OpenAI, Anthropic, and Google. 
It addresses a key limitation of large language models (LLMs) -- their reliance on static pretraining data -- by enabling standardized, dynamic access to real-time tools and environments. 
This transforms LLMs from isolated reasoning engines into adaptive, context-aware systems.

\begin{figure}[t]
\centering
\includegraphics[width=0.49\textwidth]{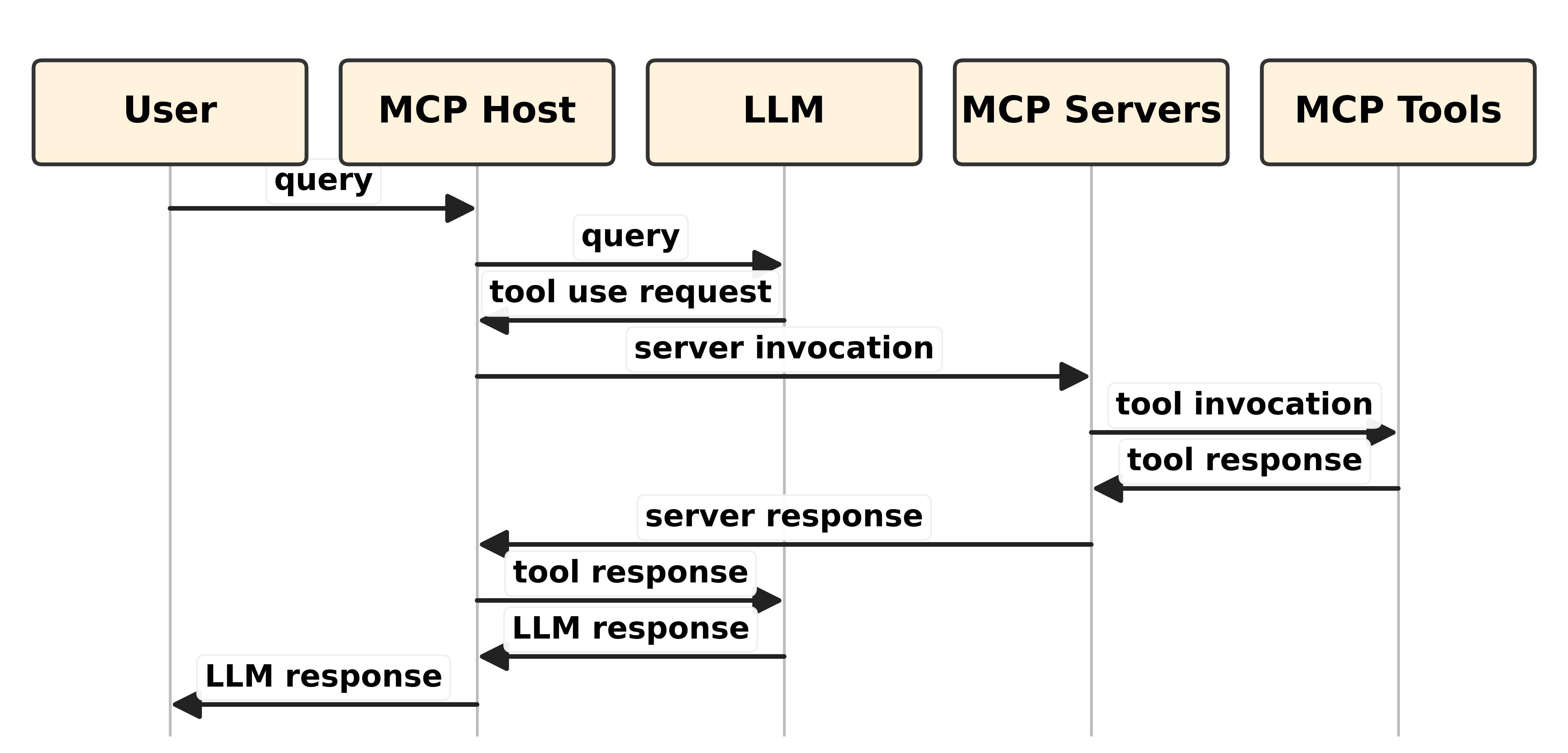}
\caption{Illustration of the MCP architecture: hosts orchestrate tool execution through distributed MCP servers.}
\label{fig:mcp_flow}
\end{figure}

In a typical workflow (\Cref{fig:mcp_flow}), an MCP \textit{host} (e.g., Cursor, Claude CLI) interacts with one or more remote MCP \textit{servers}, each exposing modular capabilities such as file I/O, API calls, or database access. 
This plug-and-play design supports scalable, composable agentic workflows without bespoke connectors or retraining.

By 2025, the open marketplace \texttt{MCP.so}~\cite{mcpso-marketplace} lists over \textbf{16,800 public servers} spanning domains from data analytics to cloud infrastructure. 
Its open-source, model-agnostic design and active developer community that spans industry and academia have made MCP the \textit{de facto} integration layer for agentic AI systems in both research and enterprise environments.

\subsection{MCP Host Security in the Enterprise}

While MCP hosts greatly enhance employee productivity, they also introduce new security risks when deployed at enterprise scale~\cite{guo2025systematicanalysismcpsecurity, hou2025modelcontextprotocolmcp}. 
Recent efforts such as \textit{MCP Safety Audit}~\cite{radosevich2025mcpsafetyauditllms} and \textit{MCP Guardian}~\cite{kumar2025mcpguardiansecurityfirstlayer} have proposed preliminary safeguards, yet comprehensive detection and mitigation mechanisms for large-scale, adaptive enterprise deployments are still lacking.
By design, MCP extends an agent’s capabilities to execute arbitrary actions, such as file access, code generation, or API calls, through external servers, many of which are third-party or community-operated. 
This expanded attack surface makes enterprise deployments susceptible to both traditional and agentic-specific threats~\cite{brett2025simplifiedsecuremcpgateways}.

MCP-based workflows pose three primary risks:
\begin{itemize}[topsep=0pt,itemsep=-1ex,partopsep=1ex,parsep=1ex,leftmargin=*]
    \item \textbf{Data exfiltration and leakage:} Sensitive enterprise data may be unintentionally or maliciously exposed through agent tool use or indirect prompt injection.
    \item \textbf{Unauthorized system access:} Misconfigured or compromised MCP servers can grant initial access, enable privilege escalation, or facilitate lateral movement across environments.
    \item \textbf{Operational disruption:} Malicious tool execution can lead to service downtime, data corruption, or resource exhaustion in production and development systems.
\end{itemize}

Detecting such risks in practice is difficult for three reasons.  
First, \textit{limited observability}: traditional security telemetry captures file and process activity but not the reasoning chains or tool invocations of AI agents, which contain the semantic context needed to distinguish malicious from benign behavior.  
Second, \textit{insufficient robustness}: static defenses constrained by pre-defined rules struggle to generalize across the diverse landscape of agentic attacks, from prompt injection to tool manipulation to credential exfiltration, while enterprise environments exhibit severe class imbalance with malicious events being extremely rare.  
Third, \textit{detection cost}: LLM-based semantic reasoning for each event is computationally expensive, requiring scalable, cost-efficient monitoring at production scale (e.g., 10,000+ daily sessions). 

These challenges motivate \ours, which integrates comprehensive observability, scalable online detection, and systematic pre-deployment red teaming to secure MCP-driven AI systems in production.

\section{\ours: System Design}
\label{sec-system}

\begin{figure}[t]
    \centering
    \includegraphics[width=0.5\textwidth]{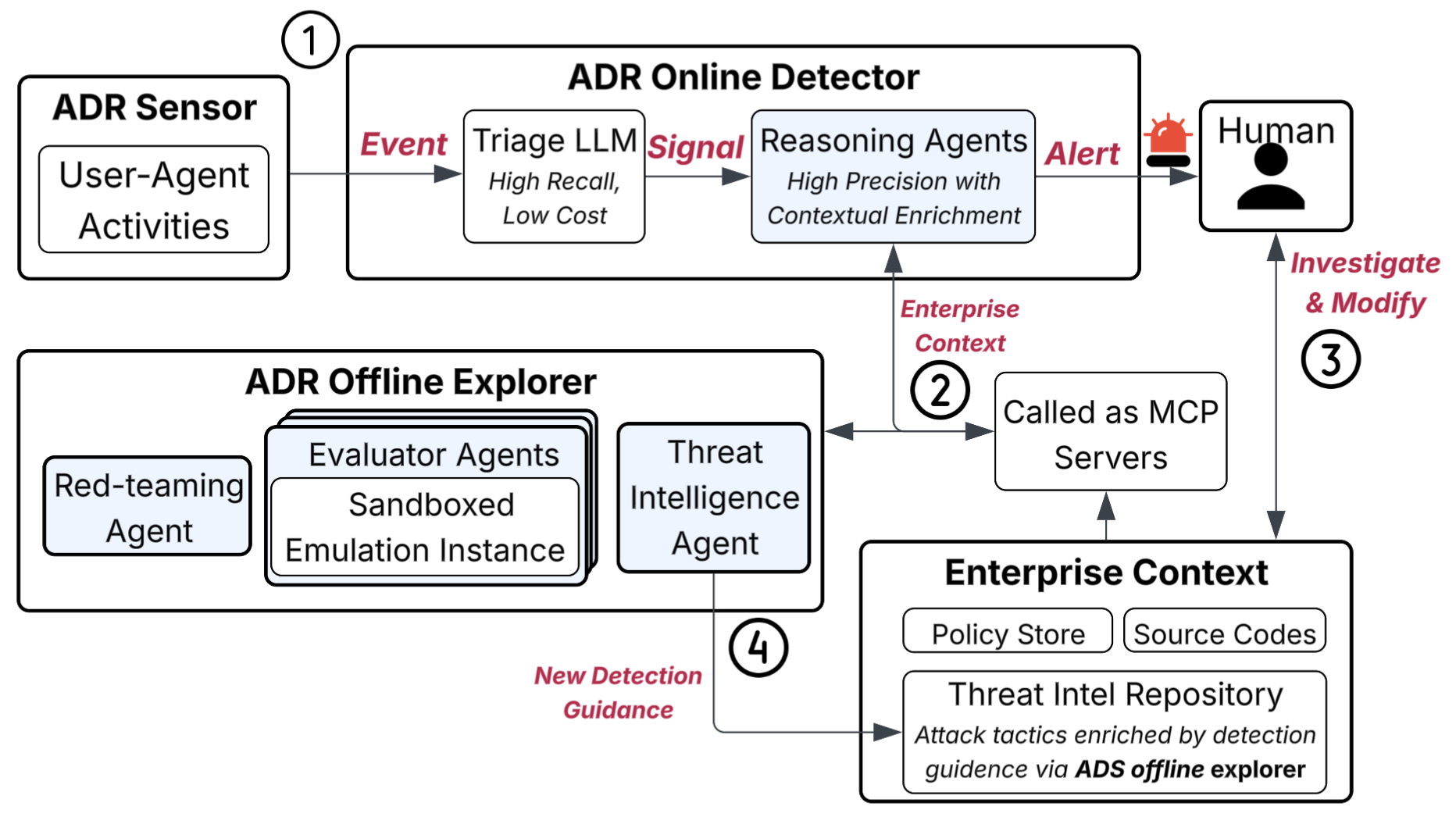}
    \caption{\textbf{Detailed \ours System Architecture.} \textbf{Left:} \ours Sensor deployment showing telemetry collection from enterprise endpoints including MCP server interactions, tool executions, and environmental context. \textbf{Center:} Two-tier online detection pipeline with Tier 1 triage processing high-volume event streams and Tier 2 analysis performing deep contextual reasoning on flagged events. \textbf{Right:} Offline EAS engine showing the evolutionary loop with a \textbf{Red-Teaming Agent} generating attack scenarios, an \textbf{Eval Agent} testing in sandboxed environments, and a \textbf{Threat Intelligence Agent} curating detection guidance into the repository.}
    \label{fig:detailed_architecture}
\end{figure}

\textbf{Core insight.}
The fundamental challenge in detecting agentic threats is the asymmetry between attackers and defenders.
Attackers exploit the semantic gap: they craft attacks that appear benign when examined superficially but are malicious when you understand the intent and context.
Traditional security tools fail because they lack two critical capabilities: (1) semantic understanding of what MCP tools actually do, and (2) enterprise-specific context about what behaviors are normal versus suspicious.

\noindent\textbf{Human operations, by design.} 
\ours mirrors how enterprise security teams work in practice (\Cref{fig:detailed_architecture}). 
The \textbf{Sensor} (\S\ref{subsec-observability}) acts like a Security Operations Center (SOC) analyst, providing comprehensive visibility into agent behavior by collecting telemetry on what agents do and why. 
The \textbf{two-tier online detector} (\S\ref{subsec-detection-loop}) mirrors SOC workflows: Tier~1 performs initial triage to catch suspicious events with high recall (minimizing missed attacks), while Tier~2 conducts deep investigation using enterprise context, similar to how detection engineers validate security incidents. This includes examining source code, consulting threat intelligence, and verifying policy compliance.
The \textbf{Offline Explorer} (\S\ref{subsec-detection-loop}) functions like an internal red team, systematically generating and testing attack scenarios in sandboxed environments during pre-deployment validation. 
Successful attacks discovered by the Explorer are curated into a threat intelligence repository that feeds back into Tier~2, strengthening the detector's robustness across diverse attack types. 
This human-inspired design makes the system adaptive and operationally grounded.

\noindent\textbf{Our approach.}
We address these challenges with three design principles (\Cref{fig:detailed_architecture}):
\textbf{(1) High-fidelity telemetry}: Collect detailed telemetry that captures not just what happened, but why it happened (the full causal chain from user prompt to agent reasoning to tool execution).
\textbf{(2) Hierarchical analysis}: Balance cost and accuracy through a two-tier architecture: fast triage flags suspicious events (\textbf{Tier 1}), then deep contextual reasoning validates true threats (\textbf{Tier 2}).
\textbf{(3) Systematic validation}: Automatically discover hard attack variants through offline red-teaming during pre-deployment testing, strengthening detection robustness (\textbf{Offline Explorer}).

We now explain each component below.

\subsection{Observability: The \ours\ Sensor}\label{subsec-observability}

\begin{figure}[t]
\centering
\includegraphics[width=0.49\textwidth]{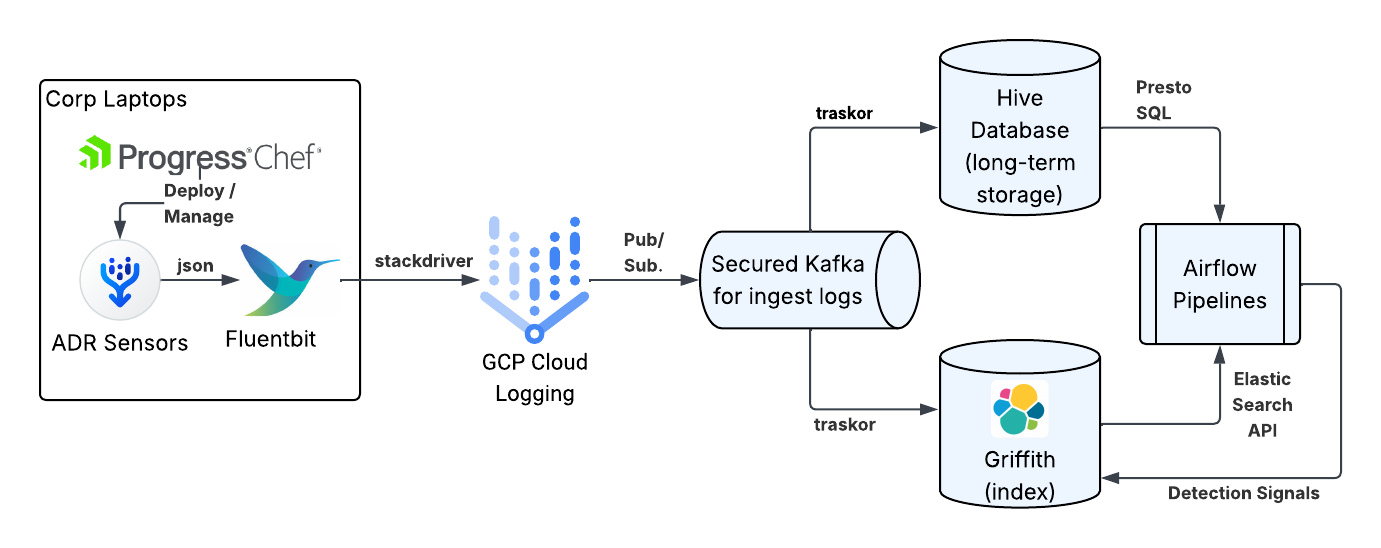}
\caption{\textbf{\ours Sensor Architecture.} The sensor parses local caches of agentic tools (Cursor, Cline, Claude Code) to reconstruct complete agent sessions, capturing user prompts, agent reasoning, MCP tool calls, and environmental context. Telemetry is forwarded to backend systems for detection analysis.}
\label{fig:sensor}
\end{figure}



\noindent\textbf{The observability gap.}
Existing Endpoint Detection and Response (EDR) tools such as were designed before the emergence of MCP-based AI workflows. They provide strong system-level visibility (e.g., file, process, and network telemetry) but lack the semantic context needed to understand \emph{why} actions occur. 
For example, when an agent writes a file, EDR sees the file write but not the user prompt that triggered it, the agent's reasoning about what to write, or which MCP tools were invoked.
This gap makes it impossible to distinguish malicious agent behavior (e.g., exfiltrating credentials) from benign actions (e.g., saving a configuration file). 

\noindent\textbf{What we collect and why it matters.}
The \textbf{\ours\ Sensor} closes this gap by reconstructing the full causal chain of agentic activity.
For each agent session, the sensor captures:

\begin{itemize}[topsep=0pt,itemsep=-1ex,partopsep=1ex,parsep=1ex,leftmargin=*]
\item \textbf{User prompts}: The original natural language instructions that initiated the agent's actions. This reveals \emph{intent} (i.e., what the user asked the agent to do).

\item \textbf{Agent reasoning}: The agent's intermediate thought process and decision-making steps. This shows \emph{how} the agent interpreted the prompt and planned its actions.

\item \textbf{MCP tool invocations}: The sequence of MCP tools called, with their arguments and execution results. This reveals \emph{what} the agent actually did and in what order.

\item \textbf{Environmental context}: MCP server configurations and installed packages (\texttt{pip}, \texttt{npm}). This provides \emph{context} about what capabilities were available to the agent.
\end{itemize}

These four dimensions together enable semantic threat detection: the detector can reason about whether an agent's actions align with user intent, whether tool usage patterns match known attacks, and whether the agent violated security policies.
For instance, if a user asks to ``summarize this Jira ticket'' but the agent reads SSH keys and makes an HTTP request, the sensor telemetry exposes this deviation from expected behavior.

\noindent\textbf{How we collect it.}
The sensor operates as a lightweight endpoint agent that parses local data stores of agentic tools (\textbf{Cursor, Cline, Claude Code}) from their SQLite databases and JSONL caches.
It correlates disparate log entries to reconstruct complete agent sessions, linking prompts → reasoning → tool calls → outcomes.
The sensor runs on an hourly schedule with minimal overhead (each run takes 0.182 seconds on average), forwarding telemetry to backend systems for analysis. The overall architecture is shown in Figure~\ref{fig:sensor}.

An alternative approach is \textbf{an LLM/MCP gateway}, which intercepts agent-to-LLM-API and agent-to-MCP-tool traffic at the network boundary. While easier to implement, gateway-based solutions require changes to MCP hosts, are incompatible with streaming responses, and capture only partial information, omitting environmental and contextual data critical for high-fidelity observability.
\cameraready{To enable effective gateway-first detection, MCP hosts would need to surface additional context/intent (e.g., the originating prompt and reasoning context) alongside tool calls (e.g., via a ``Context/Intent'' field in the MCP schema).}
\cameraready{Regarding prevention, we employ a hybrid model: the sensor enables deep forensics, while inline hooks (\S\ref{subsec:credential_exposure}) provide real-time blocking for high-severity credential leakage.}

\subsection{Detection at Scale: Online-Offline Loop}
\label{subsec-detection-loop}

Enterprise agentic traffic is heavily skewed toward benign activity, yet sophisticated attacks are designed to blend in. 
Analyzing every event with expensive LLM-based reasoning is impractical at scale; simple rule-based filtering alone misses sophisticated attacks. 
Our design combines a fast \emph{triage} pass (high recall, low cost) with selective deep \emph{reasoning} (high precision, higher cost) backed by enterprise context, strengthened through systematic offline red-teaming that discovers hard attack variants during pre-deployment validation.

\noindent
\textbf{Tier~1: Triage (initial screening).} 
Every event is first screened using a lightweight LLM-based triage prompt designed for high recall.
The triage layer flags suspicious signals such as prompt-injection phrases, requests touching credentials or permissions, role/privilege changes, and risky combinations of otherwise benign steps.
The design philosophy is conservative: when in doubt, escalate to Tier~2. Only clearly benign activity is short-circuited, ensuring minimal false negatives.

\noindent
\textbf{Tier~2: Reasoning with MCP context.} 
Events flagged by Tier~1 undergo deep semantic analysis using a reasoning agent that dynamically queries enterprise context.
The reasoning agent has access to three MCP providers supplying trustworthy enterprise information:
(1) \textbf{Source code inspection} (\texttt{get\_source\_code}) reveals what an MCP tool actually does by examining its implementation;
(2) \textbf{Threat intelligence lookup} (\texttt{get\_threat\_framework}) provides known attack tactics and techniques to ground the analysis;
(3) \textbf{Policy verification} (\texttt{get\_policies}, \texttt{assess\_policy\_violations}) checks whether actions violate organizational security rules.
The reasoning agent adaptively chooses which providers to invoke based on the event characteristics (e.g., a file write may trigger all three; a simple read may only check policies).
Tier~2 also consults curated threat intelligence from the repository (Appendix~\ref{sec:threat_repository_schema}), where each entry includes detection guidance tagged \texttt{[EAS]} (discovered by Explorer) or \texttt{[CURATED]} (manually added by security analysts).

\cameraready{\noindent\textbf{Robustness.}
Risks to robustness include brittle prompts, model updates/drift, and adversarial adaptation (e.g., attackers attempting to influence triage/reasoning).
We mitigate these risks with prompt/version pinning, change-control, and regression suites over \ours-Bench and AgentDojo before model/prompt updates.}

\noindent
\textbf{Offline red-teaming (systematic attack discovery).} 
To strengthen detection robustness across diverse attack types, the Offline Explorer systematically generates and tests attack variants through three collaborative agents:
The \textbf{Red-Teaming Agent} proposes realistic attack variants by mutating parameters and combining techniques from a seed set of known attacks.
The \textbf{Eval Agent} executes these candidates in sandboxed (isolated) environments, measuring both attack success and detection evasion.
The \textbf{Threat Intelligence Agent} curates high-value discoveries and publishes them to the threat repository for use by Tier~2.
Concretely, the Explorer runs an \emph{evolutionary algorithm}: each attack candidate is scored using a fitness function $F = \varepsilon \times \sigma \times \tau^{\alpha}$ that combines execution depth $\varepsilon$ (how far the attack progresses), semantic naturalness $\sigma$ (how benign it appears), and impact $\tau$ (damage potential).
\cameraready{Here, $\alpha$ controls how strongly we weight impact $\tau$ relative to execution depth $\varepsilon$ and naturalness $\sigma$ in $F$ (larger $\alpha$ prioritizes higher-impact variants). In our EAS configuration, we use $\alpha = 1.2$.}
Top-scoring variants survive, generate new mutations, and are re-evaluated across rounds until convergence (guaranteed by the constraint $\rho \times \mu < 1.0$, where $\rho$ is survival rate and $\mu$ is mutations per survivor).
The emulation layer acts as the feedback oracle. Detector responses and execution logs drive the evolution toward realistic, high-impact, hard-to-detect attacks.
Discovered attack patterns update Tier~2's detection logic through the threat intelligence MCP, strengthening the detector before production deployment.

\noindent
\textbf{Closing the loop.}
When the Threat Intelligence Agent updates the repository with newly discovered attacks, these entries immediately become available to Tier~2 through the threat intelligence MCP.
The updated threat intelligence steers detection in two ways: (1) it guides which MCP providers to query (e.g., prioritizing source code inspection for tool-manipulation attacks), and (2) it informs how the reasoning agent weighs evidence when making final detection decisions.
This validation process ensures \ours is tested against diverse attack patterns before deployment.

\section{\ours-Bench}\label{sec-benchmark}
\begin{table*}[t]
\small
\centering
\caption{Comparison of various benchmarks versus \ours-Bench.}
\label{tab:benchmark_comparison}
\begin{tabular}{c|c|c|c|c|c}
\hline

\multirow{2}{*}{\textbf{Benchmark}} & \multirow{2}{*}{\textbf{Use MCP}} & \multirow{2}{*}{\textbf{\# MCP Servers}} & \multirow{2}{*}{\textbf{\# Tools}} & \multicolumn{2}{c}{\textbf{Coverage}}\\
\cline{5-6}
    & & & & \textbf{\# Tasks} & \textbf{\# Threats} \\
\hline
AgentSafetyBench~\citep{zhang2024agentsafetybench} & \multirow{6}{*}{No} & \multirow{6}{*}{--} & 1702 & 2000 & 4/17 \\
ToolEmu~\citep{ruan2023toolemu} & & & 312 & 144 & 3/17 \\
AgentDojo~\citep{debenedetti2024agentdojo} & & & 70 & 97 & 4/17 \\
AgentSecurityBench~\citep{zhang2024agentsecuritybench} & & & 420 & 50 & 5/17 \\
AgentHarm~\citep{andriushchenko2024agentharm} & & & 104 & 110 & 6/17 \\
\hline
MCP-Artifact~\citep{song2025beyond} & \multirow{3}{*}{Yes} & 11 & 19 & 9 & 3/17 \\
RAS-Eval~\citep{fu2025ras} &  & 18 & 75 & {80} & 3/17 \\
MCP-AttackBench~\citep{xing2025mcp} & & -- & -- & -- & 5/17 \\
\hline
\ours-Bench (Ours) & \textbf{Yes} & \textbf{133} & \textbf{729} & \textbf{302} & \textbf{17/17} \\
\hline
\end{tabular}
\end{table*}

\begin{figure*}[t]
    \centering
    \begin{subfigure}[t]{0.235\textwidth}
        \centering
        \includegraphics[width=\linewidth]{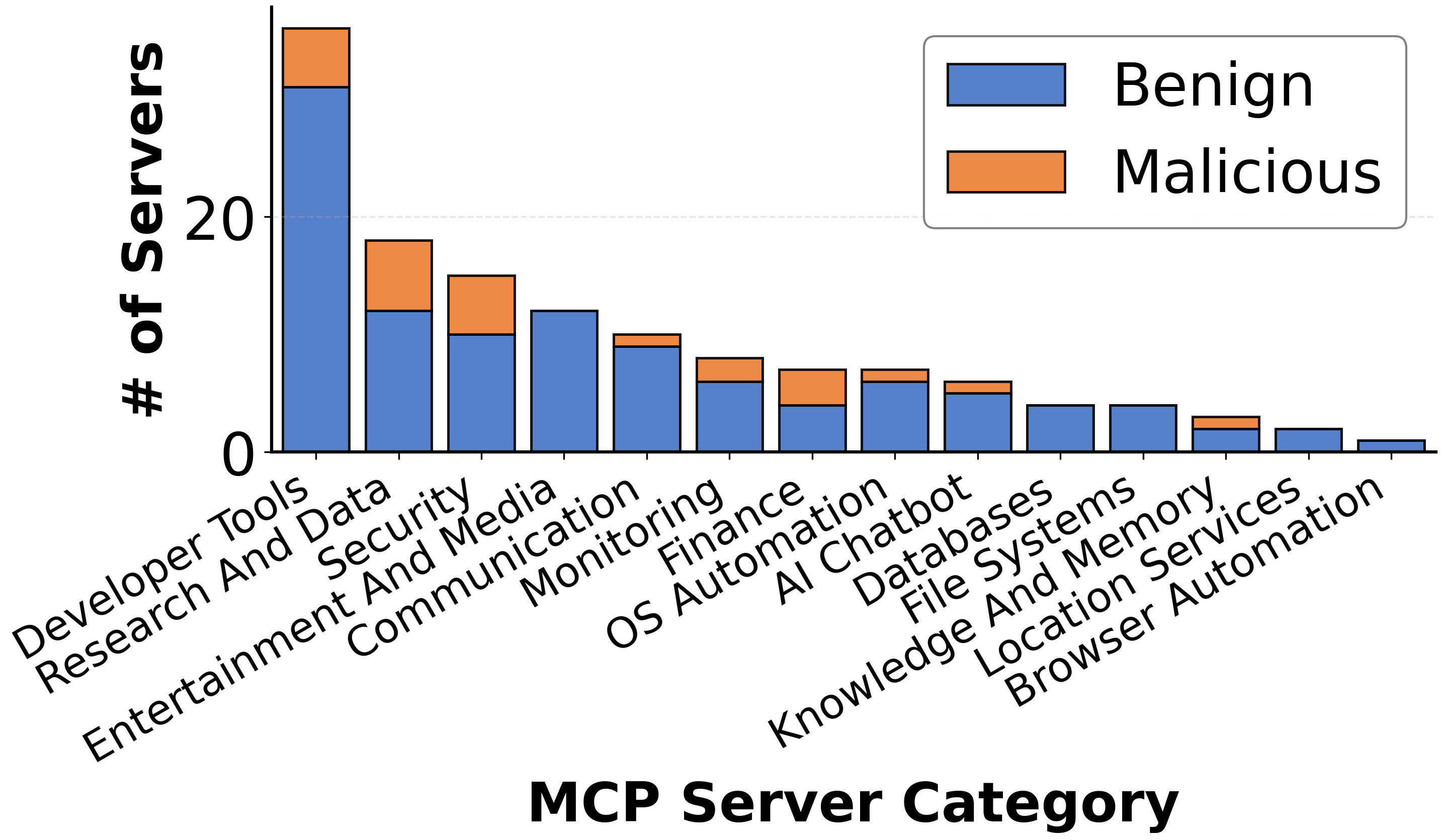}
        \caption{MCP Server Distribution}
        \label{fig:bench-server-dist}
    \end{subfigure}
    \hfill
    \begin{subfigure}[t]{0.235\textwidth}
        \centering
        \includegraphics[width=\linewidth]{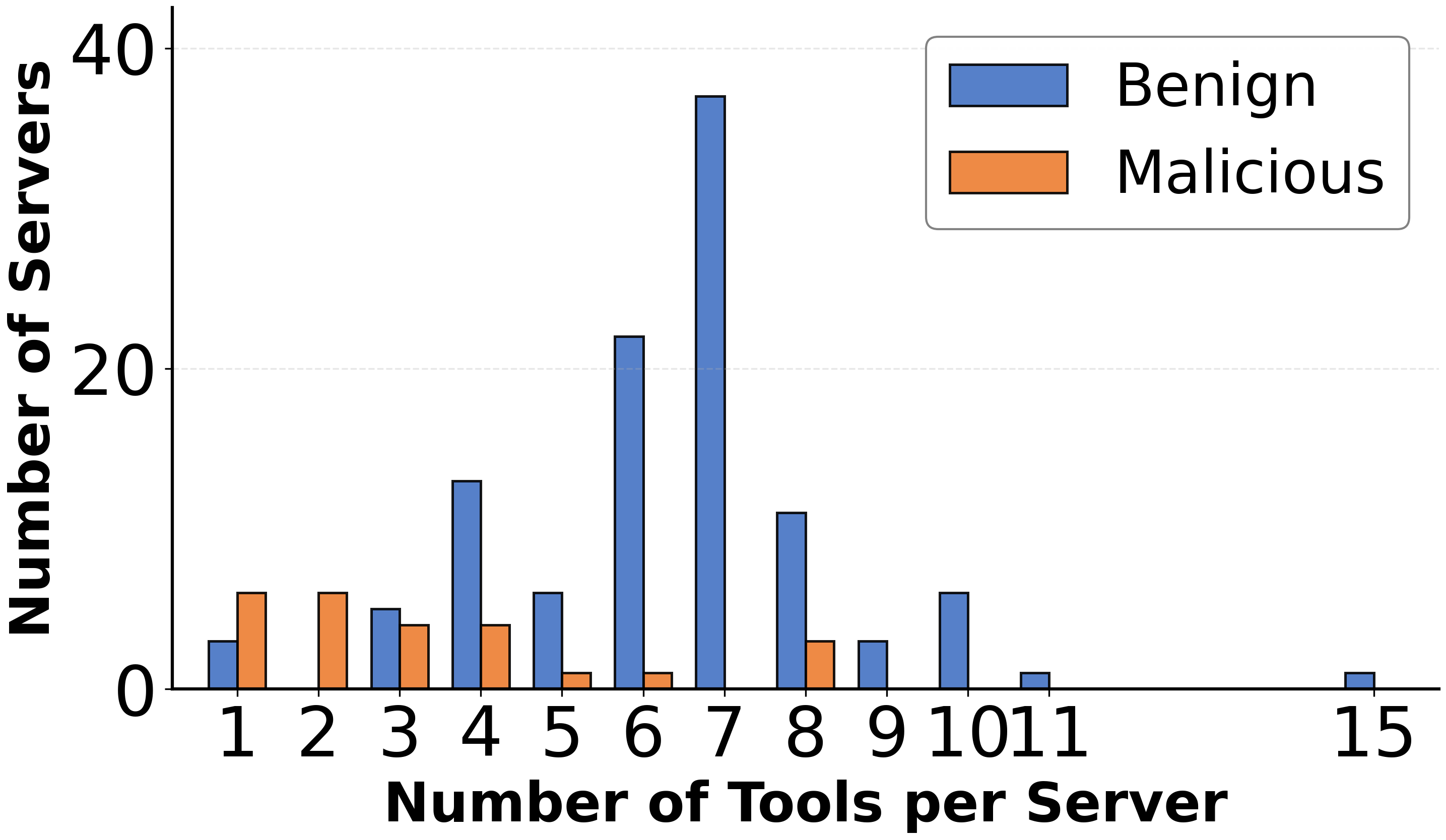}
        \caption{Tools per Server}
        \label{fig:bench-tools-dist}
    \end{subfigure}
    \hfill
    \begin{subfigure}[t]{0.235\textwidth}
        \centering
        \includegraphics[width=\linewidth]{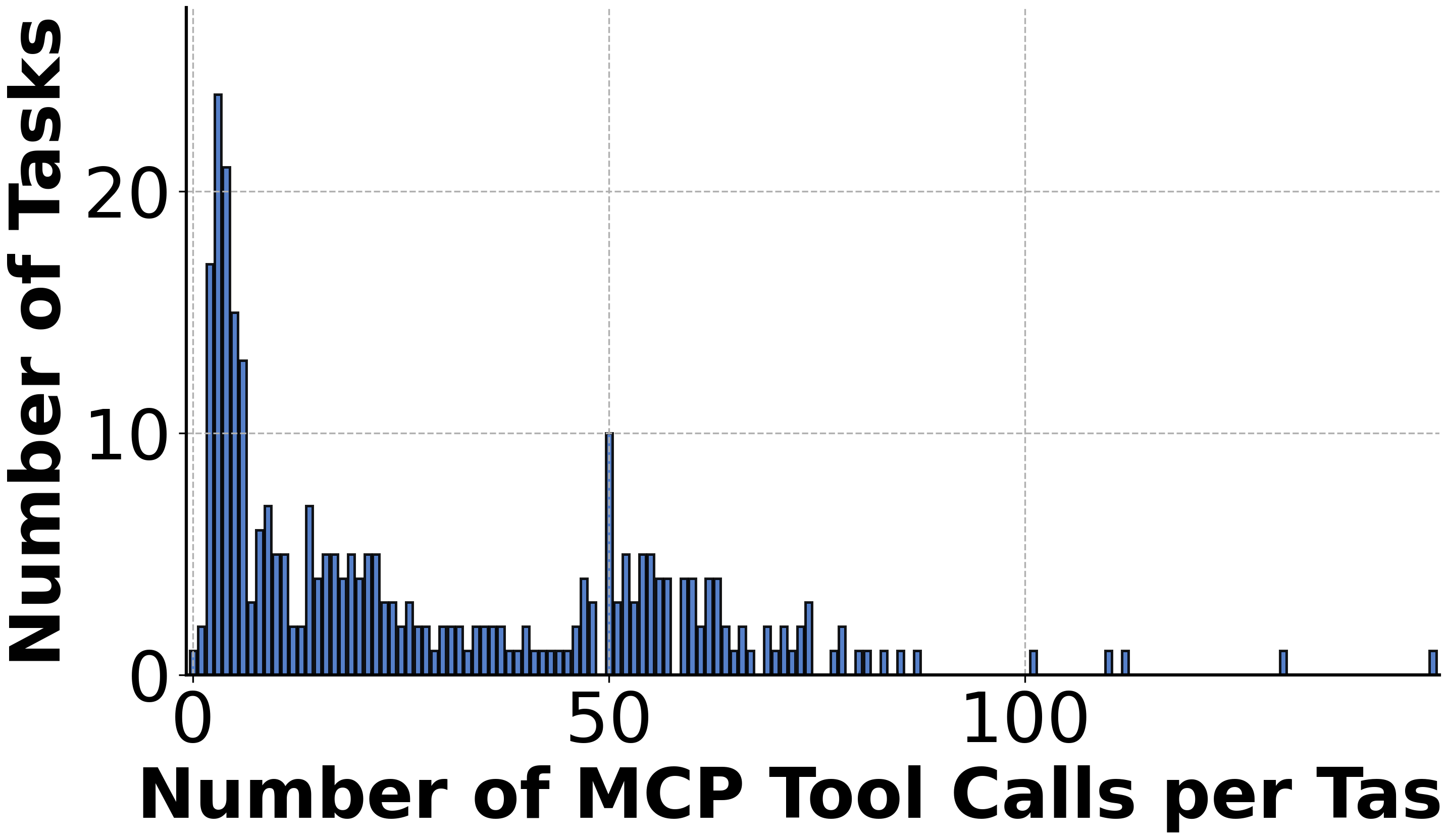}
        \caption{Tool Calls per Task}
        \label{fig:bench-task-usage}
    \end{subfigure}
    \hfill
    \begin{subfigure}[t]{0.235\textwidth}
        \centering
        \includegraphics[width=\linewidth]{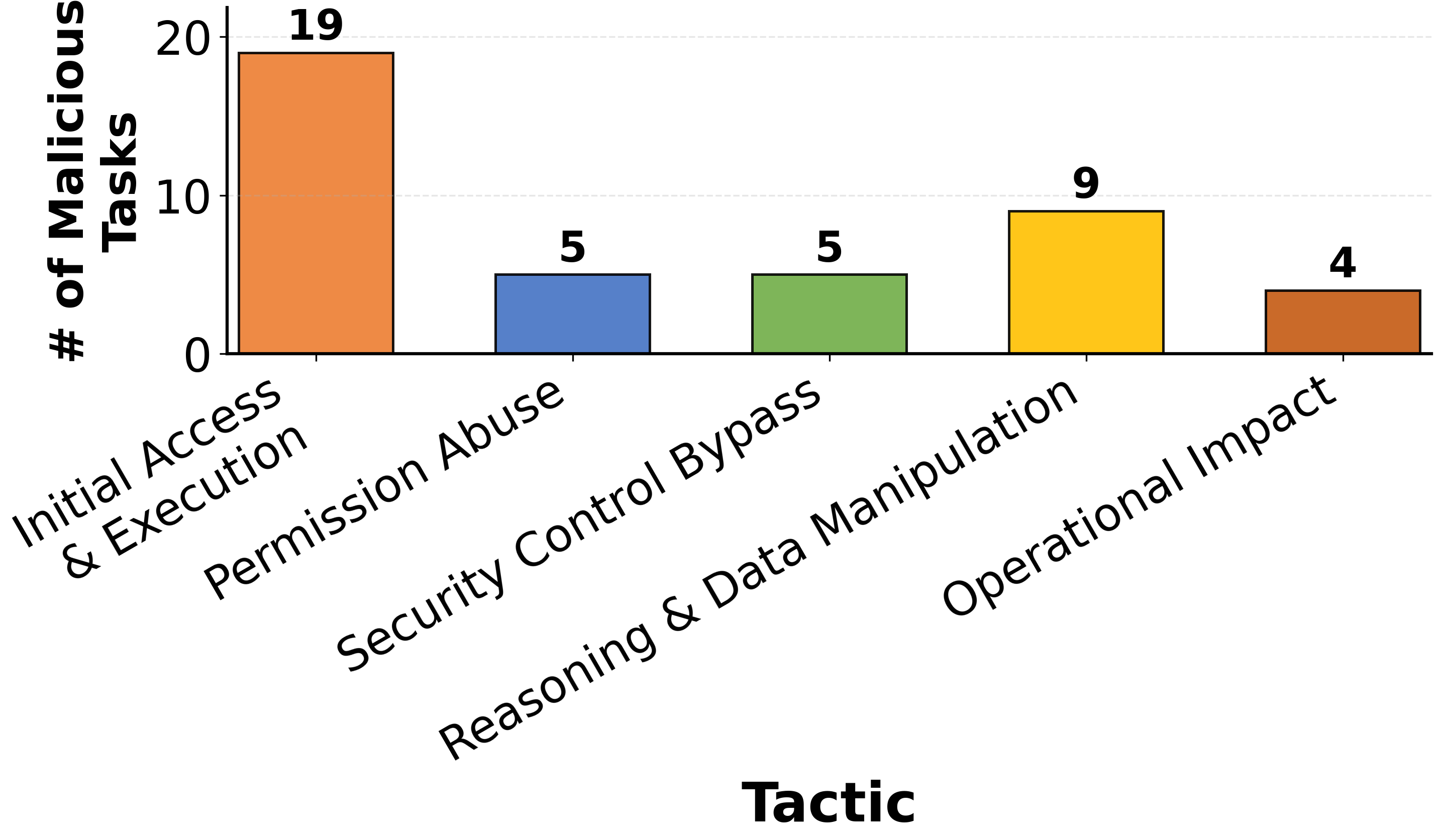}
        \caption{Malicious Tasks by Tactic}
        \label{fig:bench-tactic-dist}
    \end{subfigure}
    \caption{\textbf{Benchmark Composition and Characteristics.}
    (a) MCP servers span 14 categories with 81.2\% benign and 18.8\% malicious servers.
    (b) Benign servers provide more tools (median: 7) than malicious ones (median: 3).
    (c) Tasks invoke an average of 28.5 MCP tool calls, demonstrating realistic agentic workflows.
    (d) Malicious tasks cover 5 threat tactics with emphasis on Initial Access \& Execution.}
    \label{fig:benchmark-stats}
\end{figure*}

This section introduces \ours-Bench, derived from our enterprise experience. 
We first present the threat framework that grounds our evaluation (five tactics, 17 techniques) with evidence from public reports and enterprise telemetry. 
We then discuss limitations of existing benchmarks and contrast with prior work (\Cref{tab:benchmark_comparison}) to motivate broader threat coverage and MCP context. 
Finally, we describe \ours-Bench, including its composition, coverage, and ease of use, using summary statistics and \Cref{fig:bench-server-dist}–\ref{fig:bench-tactic-dist}.
\cameraready{\noindent\textbf{Release.}
We release \ours-Bench (task specifications, a runnable MCP registry with all MCP servers, and the evaluation pipeline) and the \ours Sensor and detection framework.
Any enterprise-specific identifiers or sensitive content are removed or replaced with safe stand-ins, enabling external researchers to reproduce our reported benchmark results and extend the benchmark.
We also release a minimal \ours configuration (JSON) aligned with our ablations, to support incremental adoption.}

\subsection{Threat Framework and Reported Instances}
\label{sec:taxonomy}
Agentic threats do not map cleanly to traditional vulnerability checklists.
We consolidate evidence from three sources: (i) public frameworks and guidelines (e.g., MITRE ATLAS~\citep{mitre_atlas_2025}, OWASP~\citep{owasp2025excessive}),
(ii) disclosed security incidents and research reports (e.g., JFrog~\citep{jfrog2025cve}, Invariant Labs~\citep{invariantlabs2025github,invariantlabs2025whatsapp,invariantlabs2025toxicflows,invariantlabs2025toolpoisoning}, Microsoft~\citep{microsoft2025mcp}, Zenity (Agent Flayer)~\citep{zenity2025agentflayer}, CyberArk~\citep{cyberark2025mcp}, Solo.io~\citep{solo2025a2a}, Trend Micro~\citep{trendmicro2025mcp}),
and (iii) operational telemetry from \company's enterprise deployment along with conversations with red team experts.
This synthesis yields a \textbf{practical five-tactic, 17-technique threat framework} tailored to MCP-driven agentic systems (Appendix~\ref{sec:threat_taxonomy}), where \emph{tactics} describe the adversary's goal (the ``why'') and \emph{techniques} describe specific attack methods (the ``how'').
The five tactics are: \textbf{Initial Access \& Execution} (6 techniques), \textbf{Permission Abuse} (2), \textbf{Security Control Bypass} (3), \textbf{Reasoning \& Data Manipulation} (4), and \textbf{Operational Impact} (2).
These techniques range from traditional attack vectors adapted for agentic systems (e.g., Indirect Prompt Injection, Tool Rug Pull) to novel agentic-specific threats (e.g., Control-Flow Hijacking, Malicious Agent Collusion).
Each technique is anchored by concrete, enterprise-style behaviors observed in the wild or reproduced in emulation.
We reference public security incidents throughout and detail how \ours-Bench reproduces them in \S\ref{subsec-bench-composition}.

\subsection{Limitations of Existing Benchmarks}
Most existing agent benchmarks cover only a small subset of agentic threats or lack MCP context. 
\Cref{tab:benchmark_comparison} shows that prior work covers only 3--6 of the 17 techniques, while \ours-Bench \textbf{covers all 17 techniques across all five tactics}. 
This comprehensive coverage with MCP context is essential for evaluating enterprise-ready detectors.

\subsection{Our Benchmark: Composition and Ease of Use}
\label{subsec-bench-composition}
To address these gaps, \ours-Bench \textbf{reproduces real-world attacks} from three sources: (i) attack scenarios adapted from existing benchmarks (e.g., MCP-Artifact~\citep{song2025beyond}, RAS-Eval~\citep{fu2025ras}); (ii) publicly reported security incidents and research disclosures; and (iii) internal threat intelligence from \company's deployment. 
We faithfully recreate patterns including indirect prompt injection (Agent Flayer~\citep{zenity2025agentflayer}), tool shadowing and MCP server manipulation~\citep{microsoft2025mcp,cyberark2025mcp}, code interpreter abuse, and supply chain compromises~\citep{jfrog2025cve}. 

\ours-Bench comprises 133 MCP servers (spanning 14 categories from file systems to cloud APIs) providing 729 distinct tools, and 302 realistic tasks averaging 28.5 tool calls per task (\Cref{fig:benchmark-stats}).
The benchmark is designed around three principles:

\begin{itemize}[topsep=0pt,itemsep=-1ex,partopsep=1ex,parsep=1ex,leftmargin=*]
    \item \textbf{Emphasis on realism}: Tasks are derived from \company's Security Operations Center (SOC) insights and enterprise deployment patterns, covering data analysis, code generation, system administration, and business process automation with authentic multi-step tool chains.
    \item \textbf{Full attack framework coverage}: Every tactic and technique in our threat framework is represented, moving beyond fragmented or narrow threat models found in prior work.
    \item \textbf{Modeling of enterprise conditions}: The benchmark mirrors challenging real-world characteristics, including highly unbalanced datasets with low attack prevalence (42 malicious out of 302 tasks, 13.9\%) and the presence of sensitive information.
\end{itemize}

\noindent\textbf{Task Sources and Enterprise Policy Store.}
We compile the 302 tasks from three sources to ensure breadth and realism.
First, we adapted tasks from existing benchmarks (e.g., MCP-Artifact, RAS-Eval) to establish comparability.
Second, we used GPT-4 to generate new benign tasks by prompting it with existing tasks and requesting innovative variants involving diverse business actions.
Third, we include 42 malicious tasks derived from the 17 attack techniques (\S\ref{sec:taxonomy}), capturing not only traditional attacks but also enterprise policy violations, where an agent's actions, while not inherently malicious, violate internal policies, providing a critical test for comprehensive security coverage.

To support such policy-centric evaluations, we compiled and generalized a set of enterprise security and compliance policies into a structured YAML store, accessible via a dedicated MCP endpoint. 
Each entry defines a canonical enterprise standard with explicit risk areas, affected roles, and enforcement conditions. 
The policy store enables consistent interpretation of policy violations across tasks, allowing for static evaluation (e.g., matching an MCP request to prohibited behaviors). 
By standardizing these enterprise norms, \ours-Bench bridges operational security controls with agentic reasoning contexts, making it suitable for evaluating detectors that must reason about compliance as well as threat behavior.

\noindent\textbf{Ease of Use and Extensibility.}
\ours-Bench is intentionally designed for modularity and rapid experimentation. 
Researchers and practitioners can extend the benchmark by following straightforward templates:
\begin{itemize}[topsep=0pt,itemsep=-1ex,partopsep=1ex,parsep=1ex,leftmargin=*]
    \item \textbf{Adding new MCP servers}: simply register additional servers in the YAML-based MCP registry, specifying entry points, authentication mechanisms, and tool definitions.
    \item \textbf{Adding new tasks}: define prompts and task configurations referencing available servers and tools; each task is a single JSON or YAML specification, making it easy to script, validate, or procedurally generate.
    \item \textbf{Adding new policies or threat techniques}: integrate additional enterprise standards or attack variants by appending them to the policy store or threat taxonomy without altering the core runtime.
\end{itemize}
All benchmark components share a uniform execution layer and deterministic emulation backend, enabling reproducibility across evaluations and straightforward CI/CD integration. 
This design allows \ours-Bench to evolve alongside the agentic security landscape, supporting future MCP protocols, toolchains, and policy domains with minimal engineering overhead.
\section{Evaluation}\label{sec-evaluation}



\subsection{Experimental Setup}\label{subsec-eval-setup}

\noindent
\textbf{Benchmarks.}
We evaluate on two benchmarks:
\begin{itemize}[topsep=0pt,itemsep=-1ex,partopsep=1ex,parsep=1ex,leftmargin=*]
    \item \textbf{AgentDojo}~\cite{debenedetti2024agentdojo} is a public benchmark for prompt injection detection containing 93 tasks (38 malicious, 55 benign) with attacks embedded in external data (tool outputs, web content, emails). \cameraready{We use AgentDojo as an auxiliary public reference focused on prompt injection. While AgentSafetyBench~\citep{zhang2024agentsafetybench} has broader coverage, it is not MCP-native, so a faithful comparison would require substantial re-instrumentation of tasks/tools into an MCP setting.}

    \item \textbf{\ours-Bench} (\S\ref{sec-benchmark}) is our enterprise threat benchmark containing 302 agentic tasks covering 17 attack techniques across 5 tactical categories. Tasks include 42 malicious scenarios (13.9\%) and 260 benign operations (86.1\%), reflecting realistic enterprise class imbalance.
\end{itemize}

\noindent\textbf{Baselines.}
We compare \ours against three detectors:
\begin{itemize}[topsep=0pt,itemsep=-1ex,partopsep=0.5ex,parsep=1ex,leftmargin=*]
    \item \textbf{LlamaFirewall}~\cite{chennabasappa2025llamafirewall} is Meta's enterprise guardrail that uses Llama Guard 3-8B for prompt classification with heuristic rules. We use default threshold settings from the official release.
    \item \textbf{GuardAgent}~\cite{xiang2024guardagent} is a policy-based system that uses GPT-4o to verify agent behavior against a knowledge graph of 47 security policies derived from OWASP guidelines and enterprise best practices.
    \item \textbf{ALRPHFS}~\cite{xiang2025alrphfs} uses hierarchical fast-and-slow reasoning: text-embedding-3-large matches agent behavior against 301 pre-trained attack patterns (fast path), and Claude Sonnet 4 analyzes ambiguous cases (slow path). We use the authors' recommended thresholds: similarity $\geq 0.8$ for harmful, $\leq 0.5$ for benign, and $0.5$--$0.8$ for slow thinking.
\end{itemize}

For all three baselines, we adapt their open-source implementations to ensure fair comparison: LlamaFirewall from Meta's Purple Llama repository~\cite{llamafirewall-codebase}, GuardAgent~\cite{guardagent-codebase} and ALRPHFS~\cite{ALRPHFS-codebase} from the authors' official releases.

\noindent\textbf{Implementation.}
\ours uses GPT-4o for triage and Claude Sonnet 4 for reasoning, with three MCP servers providing enterprise context (\S\ref{sec-system}).
Importantly, \ours requires \textbf{no hyperparameter tuning} as the triage and reasoning prompts are fixed across all tasks and benchmarks.
All detectors are evaluated under identical conditions on the same hardware.

\noindent\textbf{Metrics.}
For detection accuracy, we measure Precision, Recall, F1-score, True Positive (TP) / False Positive (FP) counts, and False Positive Rate (FPR).
For operational efficiency, we measure cost per task (\$), mean latency (seconds), cost per true positive (\$), and 95th percentile latency (seconds).

\subsection{Overall Performance}\label{subsec-eval-overall}
We benchmark \ours against all baselines across AgentDojo and \ours-Bench.

\begin{table*}[t]
\small
\centering
\caption{Performance and Cost Comparison on Two Benchmarks}
\label{tab:comparison_both}
\begin{tabular}{lccccccccc}
\toprule
\multirow{2}{*}{\textbf{Detector}} & \multicolumn{5}{c}{\textbf{Performance}} & \multicolumn{4}{c}{\textbf{Cost \& Efficiency}} \\
\cmidrule(lr){2-6} \cmidrule(lr){7-10}
& \textbf{Prec.} & \textbf{Recall} & \textbf{F1} & \textbf{TP/FP} & \textbf{FPR} & \textbf{Cost/Task (\$)} & \textbf{Latency (s)} & \textbf{Cost/TP (\$)} & \textbf{P95 Lat. (s)} \\
\midrule
\multicolumn{10}{l}{\textit{\ours-Bench (302 tasks: 42 malicious, 260 benign)}} \\
\ours (Ours)      & \textbf{1.000} & \textbf{0.667} & \textbf{0.800} & 28/\textbf{0} & \textbf{0.000} & 0.024 & 18.5 & 0.260 & 37.9 \\
ALRPHFS         & 0.333 & 0.405 & 0.366 & 17/34 & 0.130 & 0.001 & 9.2 & 0.022 & 14.8 \\
GuardAgent      & 0.231 & 0.214 & 0.222 & 9/30 & 0.115 & 0.012 & 9.4 & 0.407 & 12.5 \\
LlamaFirewall   & 0.167 & 0.190 & 0.178 & 8/40 & 0.153 & 0.459 & 93.8 & 17.377 & 225.7 \\
\midrule
\multicolumn{10}{l}{\textit{AgentDojo (93 tasks: 38 malicious, 55 benign)}} \\
\ours (Ours)      & \textbf{0.927} & \textbf{1.000} & \textbf{0.962} & \textbf{38}/3 & \textbf{0.055} & 0.012 & 18.5 & 0.029 & 47.3 \\
ALRPHFS         & 0.914 & 0.842 & 0.877 & 32/3 & \textbf{0.055} & \textbf{0.004} & \textbf{13.6} & \textbf{0.011} & \textbf{17.0} \\
GuardAgent      & 0.771 & 0.711 & 0.740 & 27/8 & 0.145 & 0.010 & 10.1 & 0.033 & 13.9 \\
LlamaFirewall   & 0.638 & 0.974 & 0.771 & 37/21 & 0.382 & 0.012 & 11.4 & 0.030 & 23.0 \\
\bottomrule
\end{tabular}
\begin{tablenotes}
\small
\item \footnotesize Bold indicates best performance. FPR = False Positive Rate, TP = True Positives, P95 Lat. = 95th percentile latency.
\end{tablenotes}
\end{table*}

\begin{figure*}[t]
    \centering
    \begin{subfigure}[t]{0.245\textwidth}
        \centering
        \includegraphics[width=\linewidth]{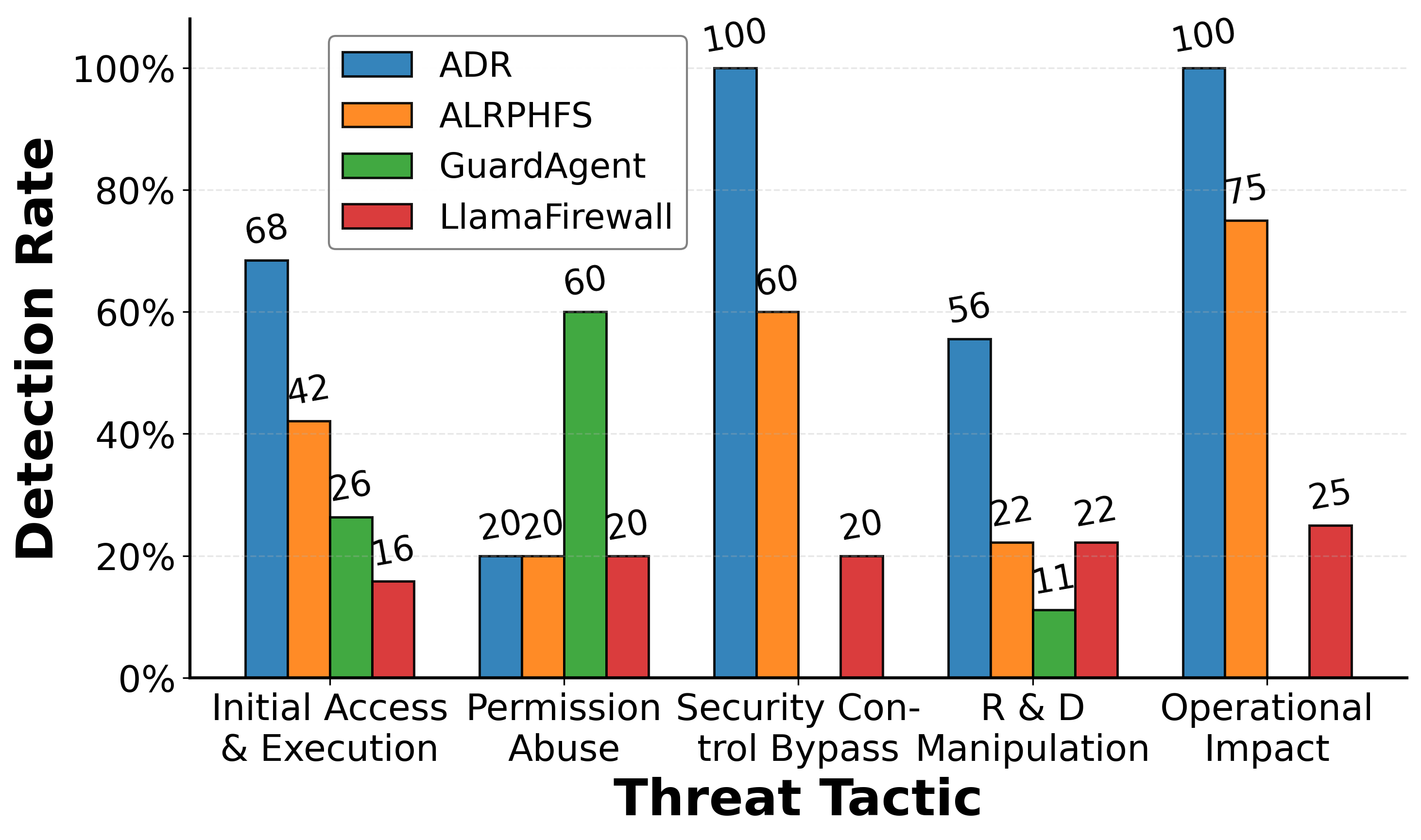}
        \caption{Detection Rate by Tactic}
        \label{fig:perf-threat-bar}
    \end{subfigure}
    \hfill
    \begin{subfigure}[t]{0.245\textwidth}
        \centering
        \includegraphics[width=\linewidth]{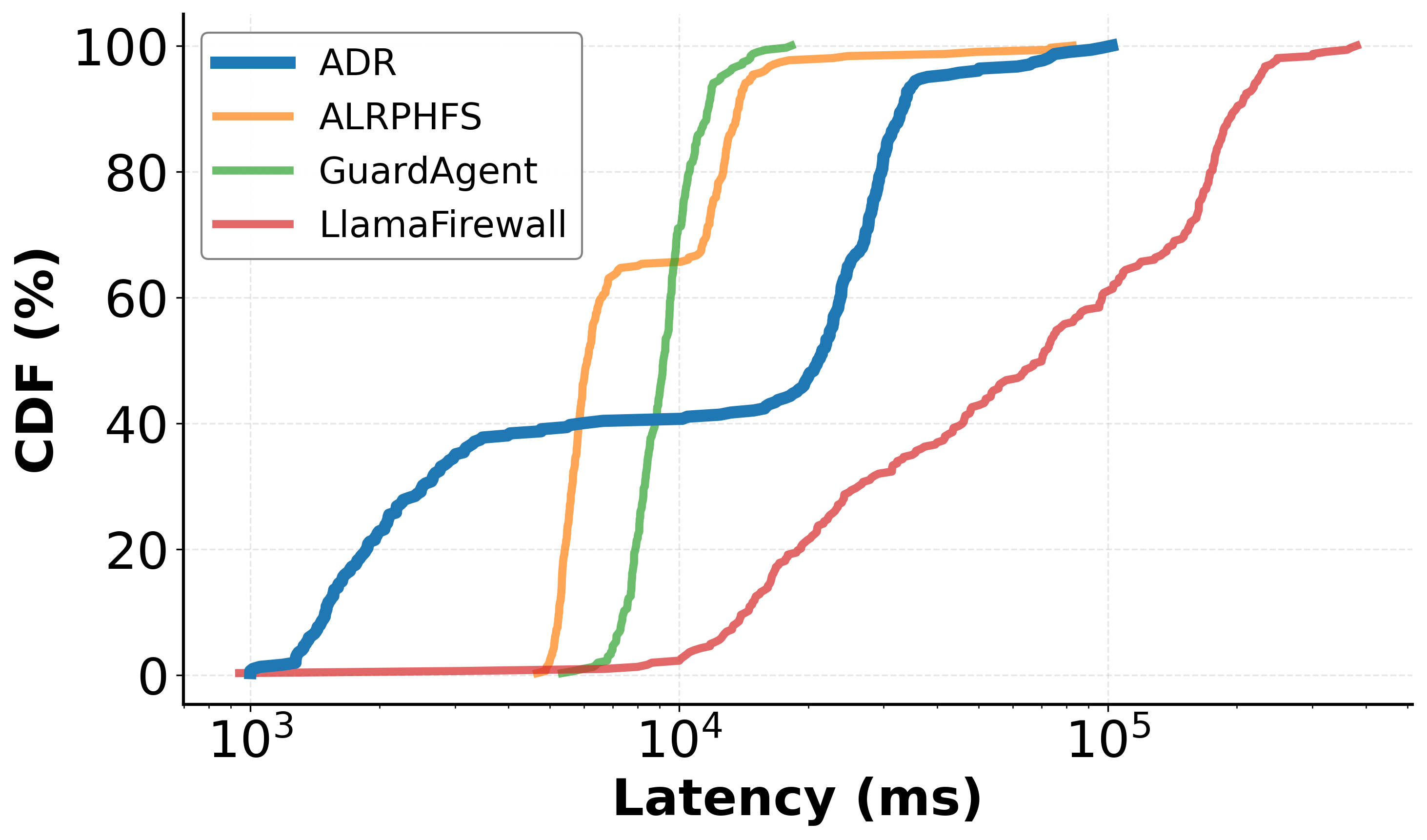}
        \caption{Latency CDF}
        \label{fig:perf-latency}
    \end{subfigure}
     \hfill
    \begin{subfigure}[t]{0.245\textwidth}
        \centering
        \includegraphics[width=\linewidth]{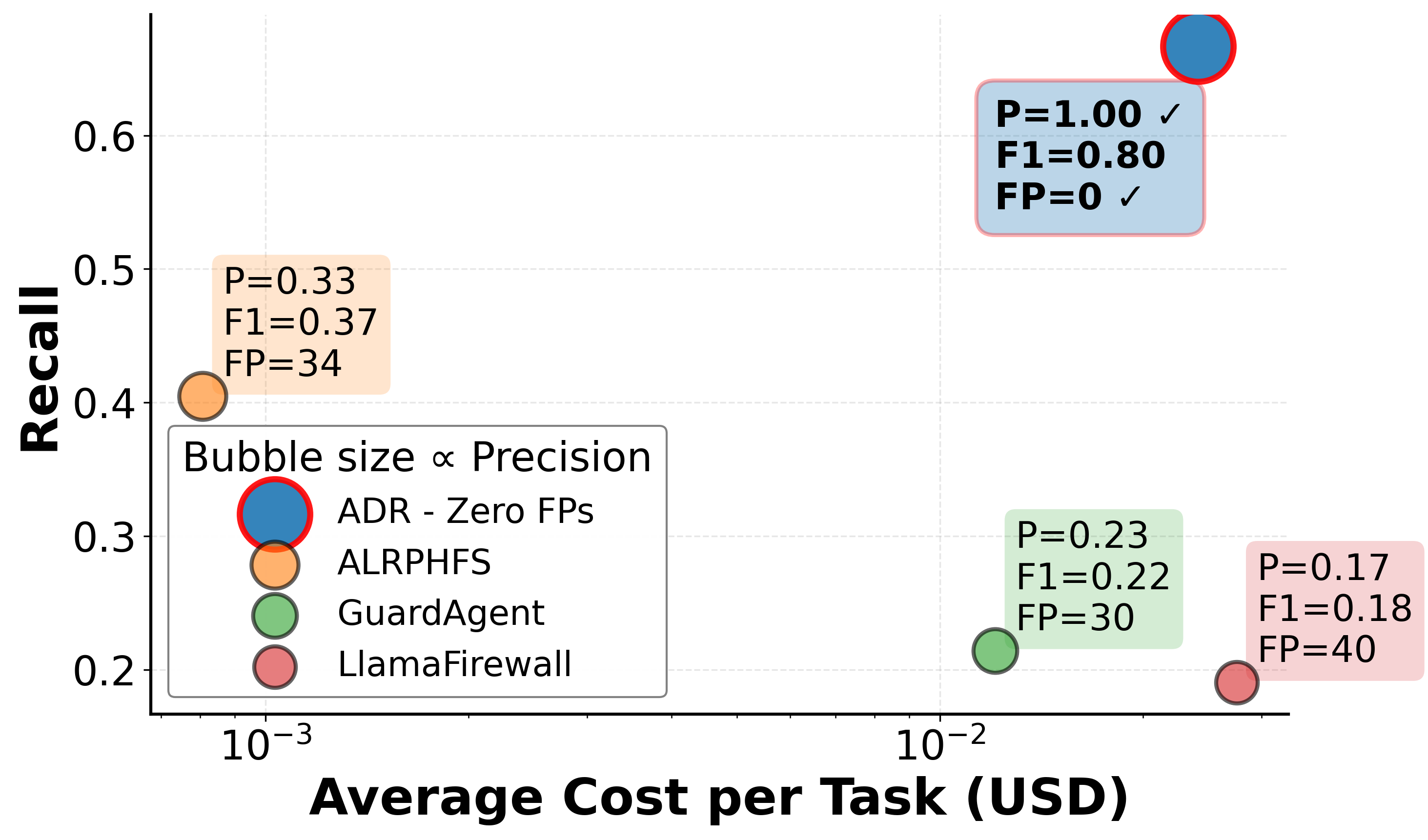}
        \caption{Cost--Recall Trade-off}
        \label{fig:perf-cost-recall}
    \end{subfigure}
    \hfill
    \begin{subfigure}[t]{0.245\textwidth}
        \centering
        \includegraphics[width=\linewidth]{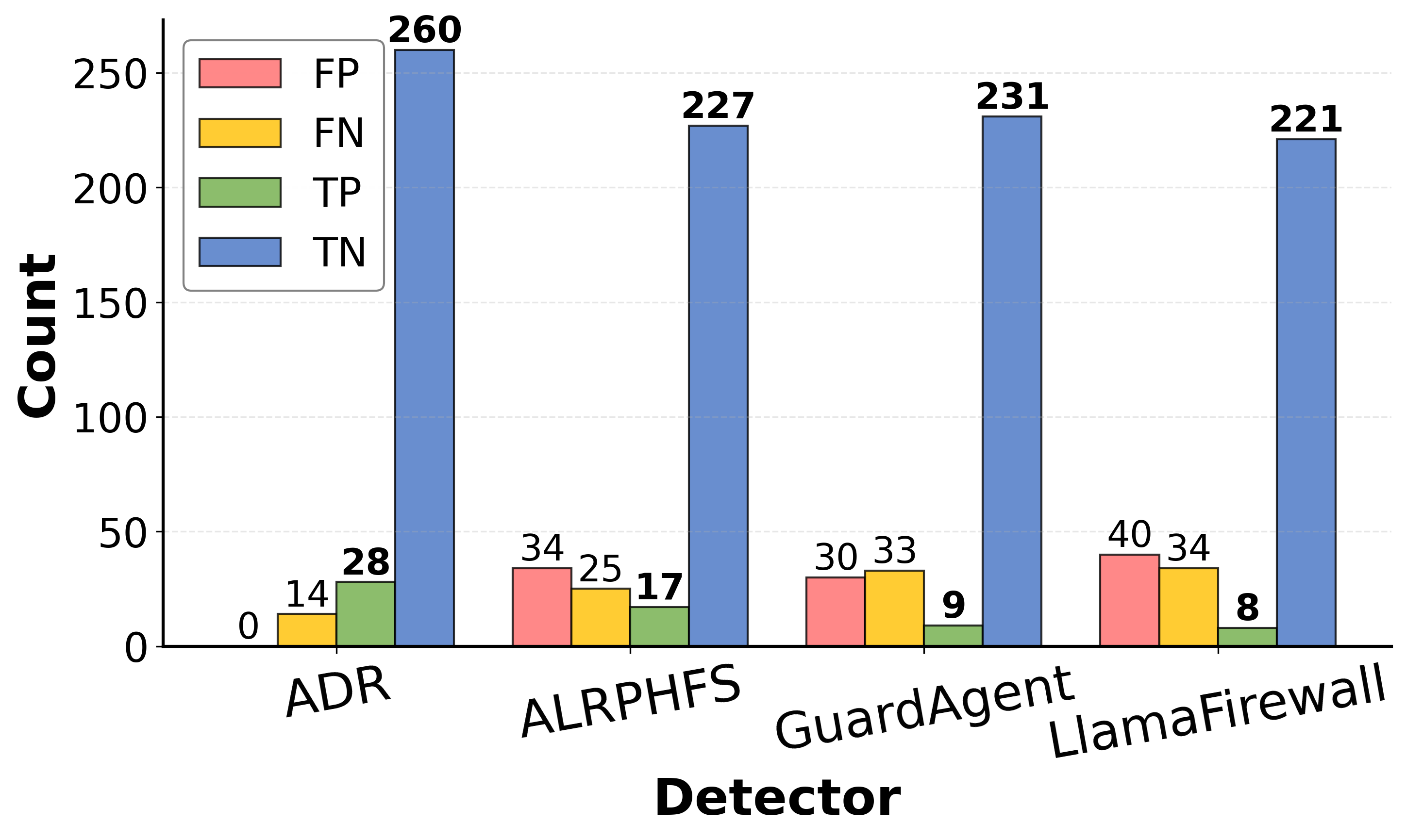}
        \caption{Confusion Matrix Comparison}
        \label{fig:perf-confusion}
    \end{subfigure}
    \caption{\textbf{Overall Performance on \ours-Bench.}
    (a) Detection rates by threat tactic show \ours's strong coverage across attack types.
    (b) Latency CDF demonstrates consistently lower tail latency.
    (c) Cost--Recall trade-off illustrates flexible operating points via triage.
    (d) Confusion matrix metrics highlight \ours's balanced precision and recall.}
    \label{fig:overall-4up}
\end{figure*}

\noindent\textbf{\ours-Bench results.}
On our enterprise benchmark (Table~\ref{tab:comparison_both}), \ours achieves \textbf{perfect precision (1.000)} with \textbf{zero false positives}, detecting 28 of 42 attacks for 0.667 recall and 0.800 F1-score.
In contrast, all baselines suffer from high false positive rates: ALRPHFS flags 34 benign tasks, GuardAgent flags 30, and LlamaFirewall flags 40.
Their precision drops to 0.333, 0.231, and 0.167 respectively, making them unsuitable for production deployment where false alarms trigger expensive incident response.

While \ours achieves moderate recall (0.667), it substantially outperforms baselines in F1-score: 0.800 versus 0.366 (ALRPHFS), 0.222 (GuardAgent), and 0.178 (LlamaFirewall).
The key insight is that \ours handles the severe class imbalance (13.9\% attack rate) through its hierarchical design: the triage layer filters obvious benign cases at low cost, while the reasoning agent with MCP context providers achieves high precision on suspicious events.

\Cref{fig:perf-threat-bar} shows varied detection rates across the 5 threat tactics: Initial Access \& Execution (68\%, 13/19 attacks), Permission Abuse (20\%, 1/5), Security Control Bypass (100\%, 5/5), Reasoning \& Data Manipulation (56\%, 5/9), and Operational Impact (100\%, 4/4), demonstrating broad coverage with particularly strong performance on control bypass and operational impact.

\noindent\textbf{AgentDojo results.}
On prompt injection attacks (Table~\ref{tab:comparison_both}), \ours demonstrates complementary strengths: \textbf{perfect recall (1.000)} detecting all 38 malicious tasks with 0.927 precision (only 3 false positives), yielding the best F1-score (0.962).
ALRPHFS achieves comparable precision (0.914) but misses 6 attacks (0.842 recall).
LlamaFirewall achieves high recall (0.974) but suffers from 21 false positives (0.638 precision).
GuardAgent achieves balanced but lower performance across both metrics (0.771 precision, 0.711 recall).
The complementary performance across benchmarks validates that \ours generalizes well across threat models: it prioritizes precision on enterprise tasks with severe class imbalance, and achieves perfect recall on prompt injection attacks where missing attacks is costly.

\noindent\textbf{Efficiency.}
\ours maintains acceptable operational costs across both benchmarks.
On \ours-Bench, it costs \$0.024 per task with 18.5s average latency (\Cref{fig:perf-latency}), which is 2$\times$ more expensive than GuardAgent (\$0.012, 9.4s) but 19$\times$ cheaper than LlamaFirewall (\$0.459, 93.8s).
\cameraready{This cost gap is largely due to \ours's enterprise-context analysis (MCP policy/source-code/threat-intel context and deep investigation).}
On AgentDojo, \ours costs \$0.012 per task with 18.5s latency, comparable to GuardAgent and LlamaFirewall.
The cost-recall trade-off (\Cref{fig:perf-cost-recall}) demonstrates the efficiency of our hierarchical design: the triage layer handles 40.7\% of tasks at minimal cost (\$0.017/task, 2.3s latency), routing only suspicious events to the more expensive reasoning agent (\$0.029/task, 29.7s latency).

\subsection{Ablation Studies}\label{subsec-eval-ablations}

To validate the contribution of each core component, we conduct ablation studies on \ours-Bench:
\begin{itemize}[topsep=0pt,itemsep=-1ex,partopsep=1ex,parsep=1ex,leftmargin=*]
\item \textbf{w/o Triage:} Remove the triage layer and send all events directly to the reasoning agent.
\item \textbf{w/o Source Code MCP:} Remove source code inspection from the reasoning agent's MCP context providers.
\item \textbf{w/o Threat Intel MCP:} Remove threat intelligence lookup from the reasoning agent's MCP context providers.
\item \textbf{w/o Policy MCP:} Remove policy verification from the reasoning agent's MCP context providers.
\end{itemize}

\begin{figure}[t]
\centering
\begin{subfigure}[b]{0.23\textwidth}
    \centering
    \includegraphics[width=\textwidth]{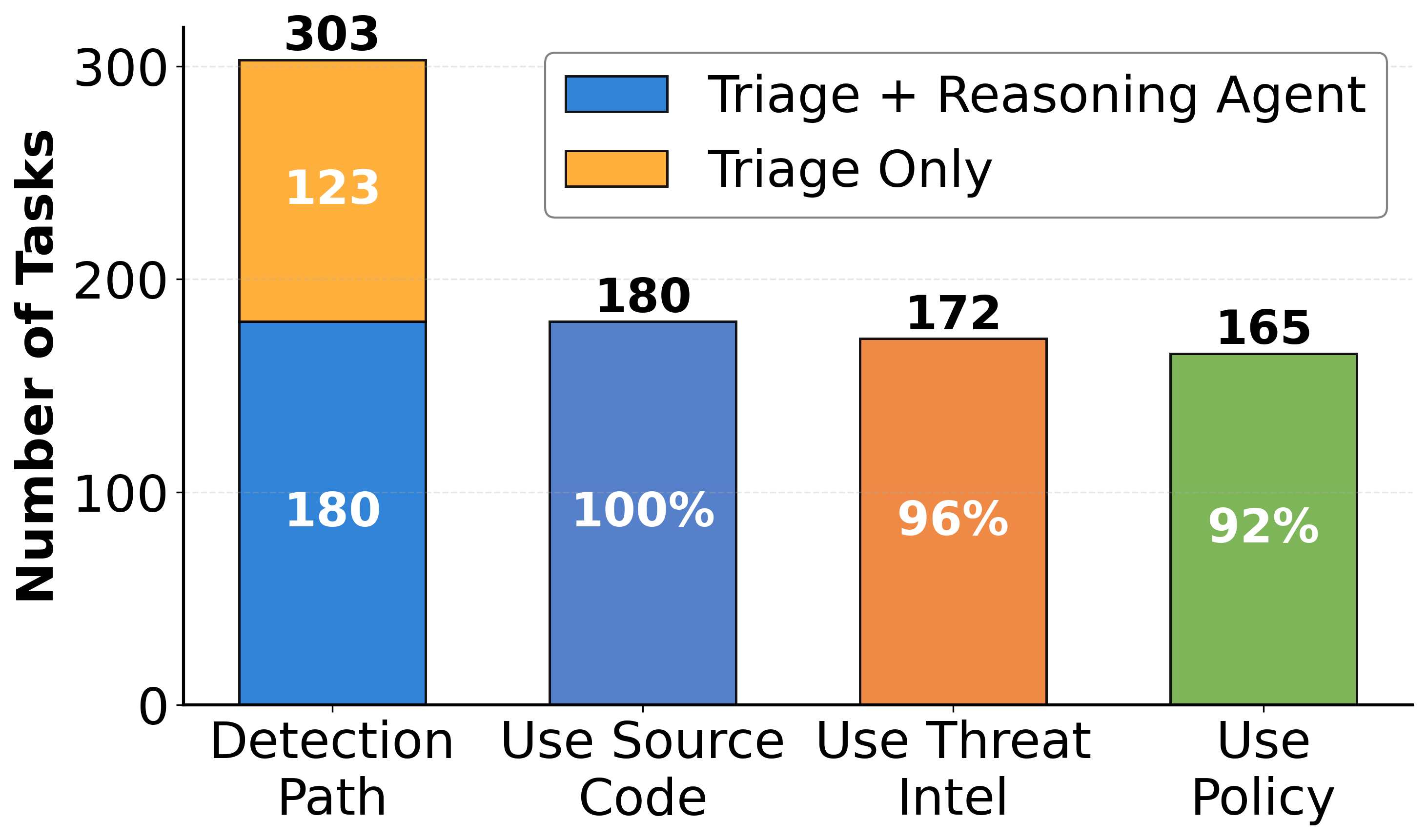}
    \caption{Detection path distribution}
    \label{fig:mcp_usage}
\end{subfigure}
\begin{subfigure}[b]{0.23\textwidth}
    \centering
    \includegraphics[width=\textwidth]{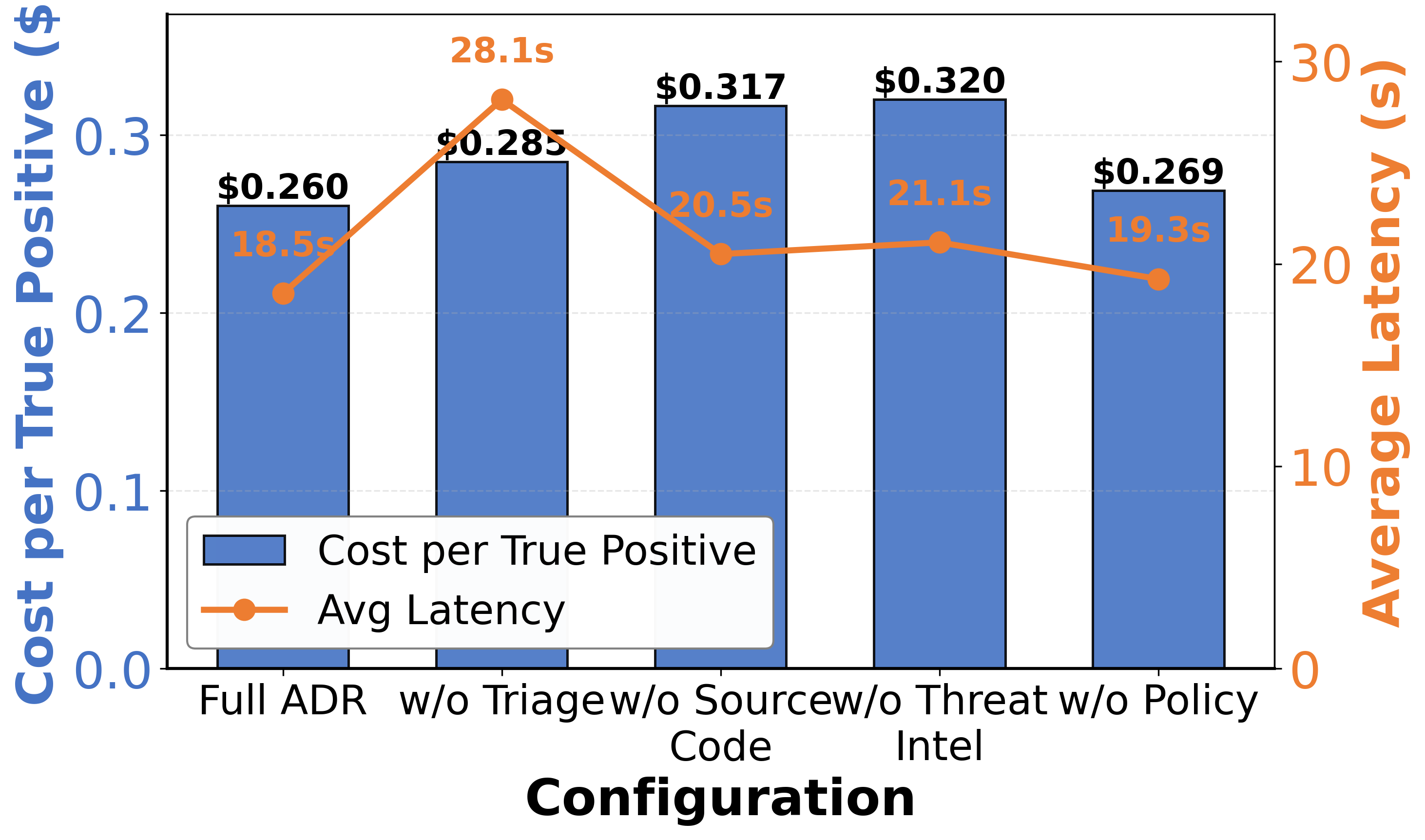}
    \caption{Cost-efficiency trade-offs}
    \label{fig:efficiency_analysis}
\end{subfigure}

\vspace{0.5cm}

\begin{subfigure}[b]{0.48\textwidth}
    \centering
    \includegraphics[width=\textwidth]{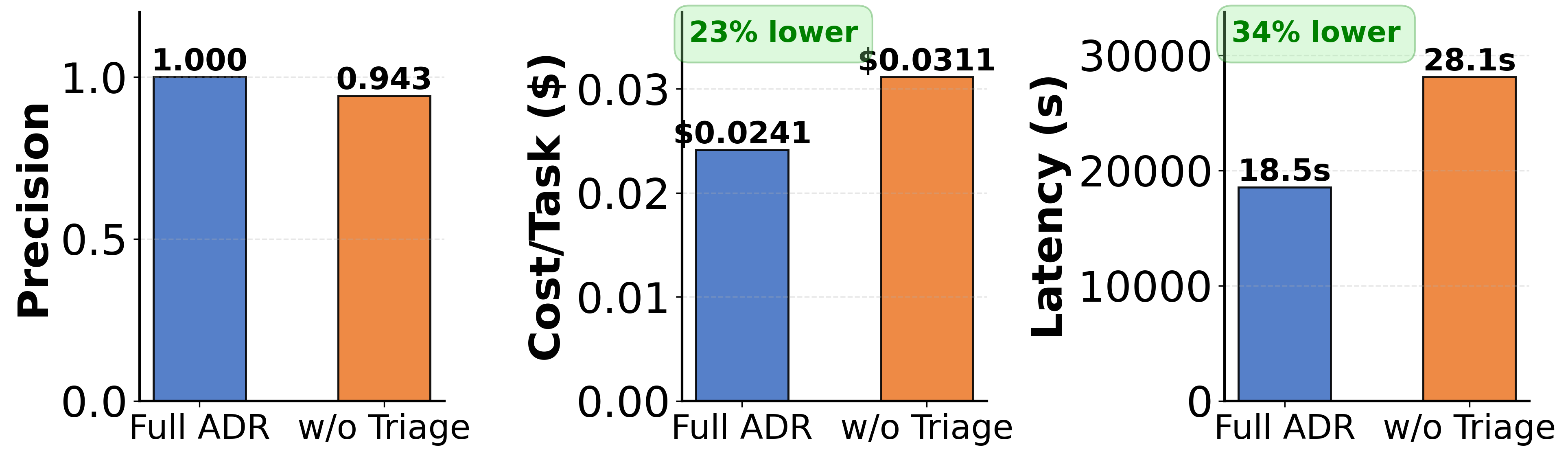}
    \caption{Triage layer benefits}
    \label{fig:triage_advantage}
\end{subfigure}

\vspace{0.5cm}

\begin{subfigure}[b]{0.48\textwidth}
    \centering
    \includegraphics[width=\textwidth]{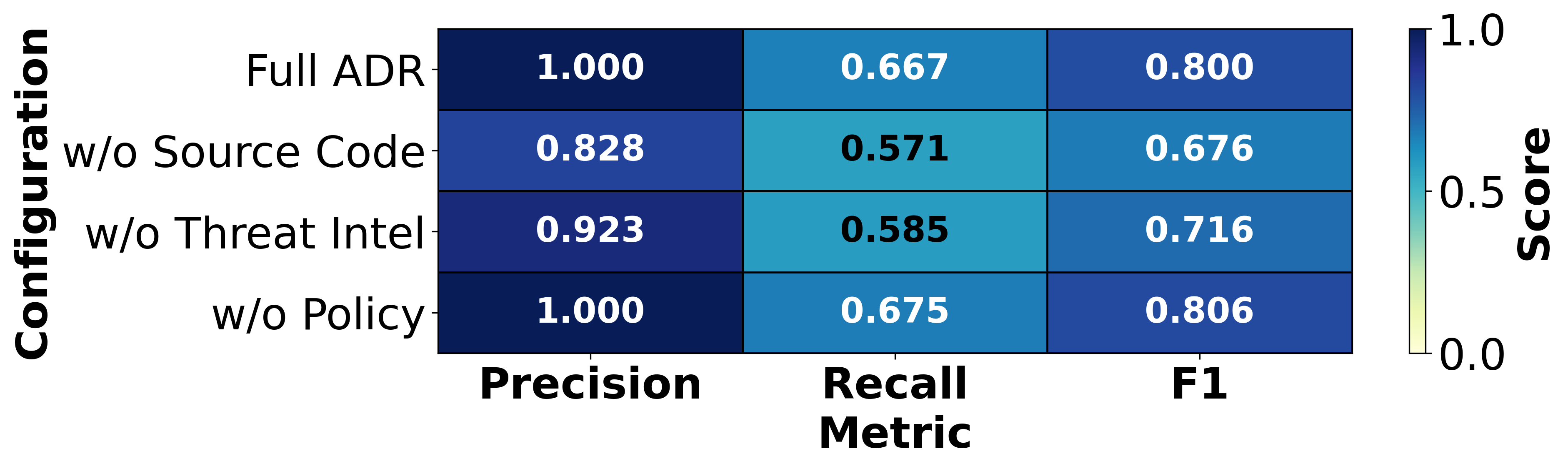}
    \caption{MCP server necessity heatmap}
    \label{fig:mcp_necessity}
\end{subfigure}
\caption{Ablation study results. (a) 40.7\% of tasks are handled by triage alone at minimal cost. (b) Cost and latency scale. (c) Removing triage introduces false positives and costs more. (d) Removing source code and threat intel MCP servers drops recall.}
\label{fig:ablation_study}
\end{figure}

\noindent\textbf{Impact of triage layer.}
Removing the triage layer increases recall from 0.667 to 0.805 and F1 from 0.800 to 0.868, but introduces 2 false positives (precision drops from 1.000 to 0.943) while increasing cost by 29\% (from \$0.024 to \$0.031 per task) and latency by 52\% (from 18.5s to 28.1s).
The triage layer handles 40.7\% of tasks at \$0.017 per task and 2.3s latency, compared to \$0.029 per task and 29.7s for the full reasoning path, validating our design: fast triage for obvious benign cases, expensive reasoning for suspicious events.
\cameraready{\noindent\textbf{Operating point selection.} Enterprises have different requirements, so there is no single universal operating point. In \ours, the primary operating-point knob is the Tier~1 routing/escalation policy: auto-resolve only clearly benign sessions and escalate uncertain/suspicious sessions to Tier~2. This directly controls the escalation rate (and thus expected cost/latency) and shifts the precision/recall trade-off; the ablation above quantifies this trade-off (recall gain vs.\ introduced FPs and higher cost/latency) when routing more traffic to Tier~2. In deployment under extreme class imbalance, we use a precision-first alerting setting to keep false alarms operationally manageable.}

\noindent\textbf{Impact of MCP context providers.}
Removing source code inspection causes the largest performance drop: recall falls from 0.667 to 0.571 (14\% relative decrease) and F1 falls from 0.800 to 0.676 (15.5\% decrease), as many enterprise attacks involve malicious code execution or tool manipulation requiring source code analysis.
Removing threat intelligence lookup also hurts performance (recall drops to 0.585, F1 to 0.716), as the reasoning agent relies on it to identify known attack patterns and tactics.
Removing policy verification has minimal impact (recall 0.667→0.675, F1 0.800→0.806), suggesting most attacks in our benchmark violate obvious security properties detectable without explicit policy checks.

\section{Real-World Deployment}\label{subsec-eval-deployment}

\begin{figure*}[t]
     \centering
     \includegraphics[width=0.75\linewidth]{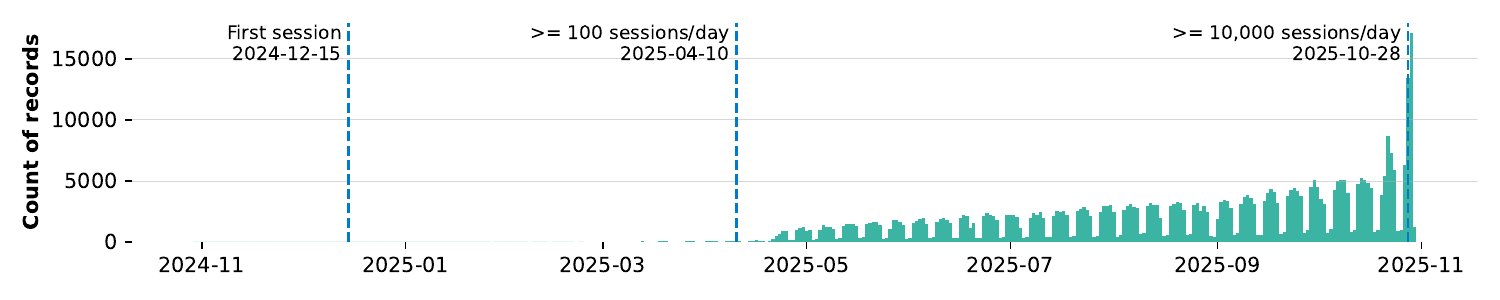}
     \vspace{-0.2in}
     \caption{Daily volume of MCP host sessions across enterprise endpoints.}
     \label{fig:deployment_daily_sessions}
 \end{figure*}

We deployed \ours across corporate MacBook endpoints (Intel and ARM) within \company. 
\cameraready{Figure~\ref{fig:deployment_daily_sessions} shows the daily volume of sessions observed in 2025.}
This large-scale production deployment provided critical insights into \ours's operational performance, coverage, and limitations under real-world enterprise workloads.

\cameraready{
    \noindent\textbf{Privacy and sensitive telemetry.}
    The system manages sensitive telemetry through a multi-layered privacy strategy that complies with the company's data minimization and strict access controls. Where possible, secrets and sensitive tokens are redacted from stored telemetry via automated scanning. Telemetry is retained for 13 months for security purposes, with access limited exclusively to security personnel under a ``least-privilege'' model. Furthermore, all data is encrypted in transit via TLS and stored in secure, auditable environments governed by enterprise security logging standards. Finally, the system has completed rigorous review by legal, privacy, and risk management teams to ensure full regulatory compliance before being deployed.
}

\subsection{Detection Summary}

\textbf{Operational outcomes.} 
\cameraready{In the deployment section, ``alerts'' are sessions routed by Tier~1/Tier~2 to a human review queue and labeled by analysts as true positives (TP, 34\%), true positives non-malicious (TPNM, 17\%), or false positives (FP, 49\%).}
The most frequent TP cases involved \textit{credential exposure}, which we further describe in \S\ref{subsec:credential_exposure}. 
TPNM cases primarily originated from internal Offensive Security teams and penetration testing activities, indicating the need to model \textit{user personas} and contextual intent. 
FPs were largely caused by context-rich sessions such as analyzing large, multi-file codebases.

\cameraready{
\noindent\textbf{Benchmark vs.\ deployment.}
\ours-Bench is designed to cover a broad set of enterprise-relevant attack tactics/techniques and benign workflows for systematic evaluation, and it is not intended to mirror the true incidence rates of different incident categories in a production environment.
In production, the real-world distribution is highly skewed (secrets/credential/data exfiltration occurs relatively often, while other attack categories are exceedingly rare), and many items labeled ``FP'' are conservative escalations on complex but benign sessions (hard-benign) that are later resolved as benign by analyst triage, rather than being ``false positives'' in a clean per-task benchmark-label sense.}

\subsection{Detecting and Preventing Credential Exposure} 
\label{subsec:credential_exposure}

\textbf{Detection.} 
\ours detected hundreds of high-severity credential exposures across 26 categories that had been inadvertently shared outside the enterprise network, posing significant security risks. 
In response, we initiated credential rotation with the owning teams and deployed preventive controls through hooks integrated into Cursor and Claude development environments.
While detection provided critical visibility, it also revealed two systemic challenges. 
\textit{First}, the sheer volume of credential-related alerts quickly exceeded the capacity for manual review, requiring an automated response mechanism. 
\textit{Second}, by the time detections were surfaced, credentials had often already been exposed externally, creating irreversible risk. 
These observations motivated a shift-left approach, moving from reactive detection to proactive prevention within the agent execution loop itself.

\textbf{Prevention.} 
Cursor and Claude Code released a feature in late 2025 called Hooks, which offers a modular mechanism to intercept and influence agent behavior within the execution loop. 
Hooks act as programmable extension points that enable external scripts to observe, modify, or block data as it traverses predefined processing stages. 
Each hook runs as a separate process, communicating with the agent via standard input/output using structured JSON messages. 
Hooks can be triggered before prompt construction, after response generation, or during reasoning steps, allowing flexible and portable integration across tools. 
Within a hook, developers can \textbf{observe} (monitor inputs, outputs, and agent state), \textbf{modify} (adjust prompts, responses, or context dynamically), or \textbf{block} (prevent disallowed or sensitive operations).
We implemented a regex-based detection mechanism using both pattern matching and entropy thresholds to identify potential secrets within prompts. 
This prevention layer is executed as a pre-prompt hook, which scans and, if necessary, blocks prompts before transmission to Cursor or Claude agents. 
\cameraready{In practice, simple non-LLM checks work well for known attacks with static patterns (e.g., secret/credential strings), but fail for attacks requiring reasoning about tool semantics, causal context, and enterprise policies.}
Our evaluation shows that this approach achieved a precision of \textbf{97.2\%}, correctly identifying \textbf{206 true positives} with only \textbf{6 false positives} across \textbf{212 unique credentials} from hundreds of thousands of MCP sessions.

\subsection{Threat Emulation of Internal/External Incidents}

\textbf{Internal Capture-the-Flag (CTF) attack}. 
\ours successfully detected simulated attacks from an internal CTF exercise. 
The attack proceeded in two stages: (1) the attacker integrated a custom shell tool into the assistant, and (2) issued a deceptive prompt instructing the agent to execute a malicious command (\texttt{curl | python3}) under the guise of sandbox testing.  
\ours detected prompt manipulation and subsequent remote code execution attempts by correlating LLM reasoning logs with MCP telemetry, generating an alert that correctly identified the misuse of MCP tools.

\begin{figure}[t]
     \centering
     \includegraphics[width=0.8\linewidth]{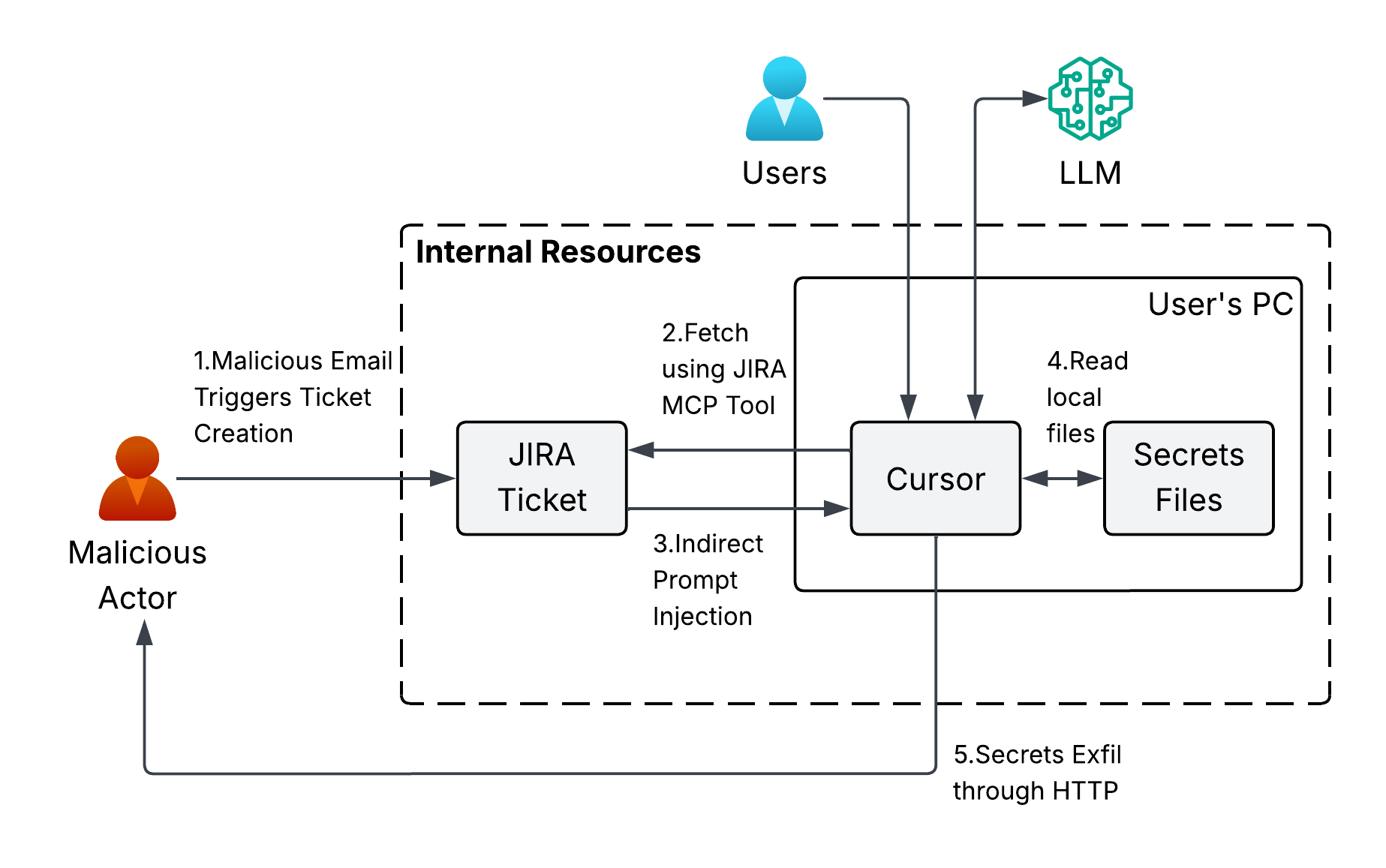}
     \vspace{-0.3in}
     \caption{Attack graph of the Agent Flayer incident illustrating indirect prompt injection via Jira–Cursor integration.}
     \label{fig:agent_flayer}
 \end{figure}

\textbf{Agent Flayer attack~\cite{zenity2025agentflayer}.}
To evaluate \ours against realistic enterprise attack patterns, we emulated high-profile industry incidents such as the Agent Flayer campaign. 
As illustrated in \Cref{fig:agent_flayer}, this threat scenario simulates an indirect prompt injection chain: a malicious email automatically creates a Jira ticket embedding hidden instructions. 
When a user's Cursor IDE -- connected via Jira's MCP integration -- retrieves and processes the ticket, the injected prompt coerces the agent into reading local configuration files and exfiltrating credentials via HTTP.
\ours accurately reconstructed and detected multiple stages of this emulated attack using fine-grained LLM and MCP telemetry. 
It flagged the prompt injection, subsequent credential access, and outbound data transfer events. 
The correlated detections demonstrated \ours's capability to trace complex multi-stage agentic threats and validate causal dependencies across tool boundaries under controlled testing conditions. 
This emulation confirmed that \ours can generalize beyond enterprise-specific patterns to detect emerging classes of real-world MCP exploitation techniques.





\section{Related Works}
\label{sec-related-works}


\noindent\textbf{Benchmarks for Agent Security.}
Several benchmarks have been developed for agentic AI security including ToolEmu~\cite{ruan2023toolemu}, AgentDojo~\cite{debenedetti2024agentdojo}, AgentHarm~\cite{andriushchenko2024agentharm}, AgentSafetyBench~\cite{zhang2024agentsafetybench}, and AgentSecurityBench~\cite{zhang2024agentsecuritybench}. 
However, they primarily focus on prompt injection and basic tool misuse and do not support native MCP, making them less relevant for the enterprise architectures that are now standard.
Recent benchmarks~\cite{song2025beyond,fu2025ras,xing2025mcp} directly address this gap by specifically modeling MCP security. 
While this is a critical step forward, they are still limited in two key dimensions: they provide incomplete coverage of the full threat landscape (\S\ref{sec-benchmark}) and lack realism by focusing exclusively on attack scenarios, failing to account for enterprise environments where agentic activity is predominantly benign. 
These gaps in coverage and realism motivate \ours-Bench.
    
\noindent\textbf{Defending LLM Agents and MCP Security.} 
Defense systems for agentic security primarily fall into two categories: preventative mechanisms that block malicious actions in real-time, and detective mechanisms that identify malicious agentic events for post-processing.
Preventative mechanisms have evolved from static, rule-based systems, such as regex and allowlisting, to more dynamic LLM-based guardrails.
State-of-the-art approaches use LLM reasoning to enforce safety policies at runtime, including industry guardrails~\cite{costa2025securing,rebedea-etal-2023-nemo,amazon_bedrock,google_guardrails,AI_Infra_Guard} such as Anthropic's constitutional AI~\cite{bai2022constitutional} and Meta's LlamaFirewall~\cite{chennabasappa2025llamafirewall}. 
In academia, GuardAgent~\cite{xiang2024guardagent} and AGrail~\cite{luo2025agrail} generate adaptive safety checks and executable code to validate agent actions against security requirements. 
ShieldAgent~\cite{chen2025shieldagent} structures policy documents into verifiable rule circuits to shield protected agents. 
While powerful, these systems are fundamentally constrained by the policies they are given and struggle to defend against unknown or emergent threat patterns.
Detective mechanisms aim to find unknown threats by moving beyond fixed rules. 
While traditional anomaly detection can be applied to agent logs, such methods often fail to interpret the semantic context of agentic workflows. 
More advanced approaches like ALRPHFS~\cite{xiang2025alrphfs} address this by using adversarial learning to extract ``risk patterns'' from adversarial interactions. 
However, all defenses rely on pre-defined rules or learned patterns and cannot proactively discover and adapt to emergent, zero-day threats in live agentic systems. 

\noindent\textbf{Automated Red Teaming.}
To discover zero-day threats, recent works use automated red teaming via techniques such as genetic algorithms~\cite{liu2023jailbreaking,lapid2024open}, gradient-based search~\cite{zou2023universal,chen2024llm}, and agent-based frameworks~\cite{xu2024redagent,wang2024ali,jiang2024wildteaming,liu2024autodan}. 
However, they remain limited to optimizing prompt designs. 
Even the most advanced framework, AutoRedTeamer~\cite{zhou2025autoredteamer}, which introduces a modular attack toolbox and a strategic memory architecture, is still designed to find sophisticated jailbreaking methods for the LLM itself.
This collective focus on foundational LLM security leaves a critical gap in understanding vulnerabilities for agentic systems. 
\ours addresses this gap by focusing on vulnerability discovery for agentic systems operating at enterprise scale.
\section{Conclusion}\label{sec-conclusion}
We presented \ours, the first enterprise-scale framework for securing AI agents operating through the Model Context Protocol.
\ours addresses three core challenges---limited observability, insufficient robustness, and high detection costs---via three integrated components:
the \ours Sensor for high-fidelity agentic telemetry,
a two-tier online Detector that balances cost and precision through hierarchical triage and context-aware reasoning,
and an offline Explorer that systematically discovers hard attack variants through evolutionary red-teaming.

On \ours-Bench (302 tasks, 133 MCP servers, 17 attack techniques), \ours achieves zero false positives while detecting 67\% of attacks, outperforming three state-of-the-art baselines by 2--4$\times$ in F1-score.
On AgentDojo, \ours detects all prompt injection attacks with only three false alarms.
Deployed at \company for over ten months across 7,200+ hosts processing 10,000+ daily sessions, \ours uncovered hundreds of credential exposures and enabled a shift-left prevention layer with 97.2\% precision.

We release \ours-Bench, the \ours Sensor, and the detection framework to support reproducibility and community adoption.
Looking ahead, we see opportunities in extending \ours to multi-agent coordination protocols, adaptive real-time prevention at the MCP gateway layer, and tighter integration with evolving MCP standards to further close the gap between detection and response.



\bibliography{reference}
\bibliographystyle{mlsys2025}

\appendix
\section{Appendix}

\subsection{Enterprise Agentic Threat Taxonomy}
\label{sec:threat_taxonomy}

Table~\ref{tab:threat_taxonomy} presents our complete five-tactic, 17-technique threat taxonomy with references to public security incidents.

\subsection{Threat Repository Schema (YAML Excerpt)}
\label{sec:threat_repository_schema}

\begin{AdsCodeYAML}
threat_framework:
  tactics:
    initial_compromise:
      techniques:
      - id: ADR.T0001
        name: Insecure Supply Chain for Agentic Components
        detection_guidance:
        - 'Malicious: Monitor for expanded permissions in config_manager accessing financial data sources...'  # [EAS]
        - 'Malicious: Monitor for unauthorized access to repository secrets and API endpoint creation using github_server...'  # [EAS]
      - id: ADR.T0002
        name: Indirect Prompt Injection
        detection_guidance:
        - 'Malicious: Monitor legal research tools for unusual document scanning and contract analysis patterns...'  # [CURATED]
\end{AdsCodeYAML}

\noindent
Each technique record contains three fields: \texttt{id}, \texttt{name}, and \texttt{detection\_guidance}.
Guidance lines are tagged \texttt{[EAS]} (discovered by the offline Explorer) or \texttt{[CURATED]} (added by analysts).
Tier~2 uses these lines to select which MCP providers to query and to weigh evidence during detection decisions (\S\ref{subsec-detection-loop}).

\clearpage
\begin{table*}[!t]
\vspace*{-38em}
\footnotesize
\centering
\caption{Enterprise Agentic Threat Taxonomy: 5 Core Tactics}
\label{tab:threat_taxonomy}
\begin{tabular}{@{}p{2.8cm}p{5.5cm}p{6.5cm}@{}}
\toprule
\textbf{Tactic} & \textbf{Description} & \textbf{Techniques \& References} \\
\midrule
\textbf{Initial Access \& Execution}
& Adversaries gain initial control and execute malicious actions within the agentic environment by exploiting vulnerabilities, injecting malicious instructions, or manipulating tool behavior
& \textbf{6 techniques:} Insecure Supply Chain \citep{jfrog2025cve}, Indirect Prompt Injection \citep{zenity2025agentflayer}, Control-Flow Hijacking \citep{invariantlabs2025toxicflows}, Code Interpreter Abuse \citep{cyberark2025mcp}, Insecure Output Handling, Tool Rug Pull \citep{invariantlabs2025toolpoisoning} \\
\midrule
\textbf{Permission Abuse}
& Adversaries exploit or exceed authorized permissions to access sensitive data and resources beyond their legitimate scope
& \textbf{2 techniques:} Exploitation of Excessive Tool Permissions \citep{invariantlabs2025github,invariantlabs2025whatsapp}, Agent Identity Spoofing \\
\midrule
\textbf{Security Control Bypass}
& Adversaries evade detection and circumvent security mechanisms by deploying malicious tools, manipulating tool resolution, or coordinating multiple agents
& \textbf{3 techniques:} Tool Shadowing \citep{microsoft2025mcp,cyberark2025mcp}, Tool Hallucination Manipulation, Malicious Agent Collusion \citep{solo2025a2a} \\
\midrule
\textbf{Reasoning \& Data Manipulation}
& Adversaries corrupt data sources or manipulate an agent's reasoning processes to compromise decision integrity and degrade long-term reliability
& \textbf{4 techniques:} Unvetted MCP Server Connection, Semantic Data Poisoning, Long-Term Goal Hijacking \citep{hubinger2024sleeper}, Temporal Data Attack \\
\midrule
\textbf{Operational Impact}
& Adversaries disrupt availability and degrade business operations by exhausting system resources or overwhelming the agent's inference capacity
& \textbf{2 techniques:} Agent-Facilitated Resource Exhaustion \citep{owasp2025excessive}, Model-Layer Denial of Service \\
\bottomrule
\end{tabular}
\end{table*}


\end{document}